\documentclass{elsarticle}

\usepackage{amsmath}

\usepackage{amssymb}

\newcommand{\e}[1]{\emph{#1}}
\newcommand{\bftab}{\fontseries{b}\selectfont}

\newcommand{\mb}[1]{\mathbb{#1}}
\newcommand{\sbeq}{\subseteq}

\newcommand{\lra}{\longrightarrow}
\newcommand{\lrma}{\longmapsto}

\DeclareMathOperator{\NN}{NN}

\newcommand{\owa}[1]{\operatorname*{\mathnormal{#1\!\downarrow}}}

\DeclareMathOperator{\lp}{lp}
\DeclareMathOperator{\alp}{alp}

\DeclareMathOperator{\lrd}{lrd}
\DeclareMathOperator{\lof}{lof}

\usepackage{amsthm}
\theoremstyle{definition}
\newtheorem{definition}{Definition}

\usepackage[font=footnotesize]{subfig}
\usepackage{etoolbox}
\robustify\subref

\usepackage{booktabs}
\usepackage{longtable}
\usepackage{siunitx}

\usepackage{tikz}
\usetikzlibrary{matrix,shapes,arrows,positioning,chains}

\usepackage[hidelinks]{hyperref}

\usepackage{fancyhdr}

\fancypagestyle{firstpagestyle}
{
\fancyhead[R]{\thepage\ifodd\value{page}\else\hfill\fi}
\fancyhf{}
\fancyhead[C]{\normalfont\scriptsize To appear in \href{https://doi.org/10.1016/j.patcog.2021.107991}{Pattern Recognition 118 (2021) 107991}}
\fancyfoot[C]{\scriptsize © 2021. This manuscript version is made available under the \href{http://creativecommons.org/licenses/by-nc-nd/4.0/}{CC-BY-NC-ND 4.0 license}.}%
}

\begin{document}
\begin{frontmatter}

\title{Average Localised Proximity: A new data descriptor with good default one-class classification performance}

\author[add1]{Oliver Urs Lenz\corref{cor1}}
\ead{oliver.lenz@ugent.be}
\cortext[cor1]{Corresponding author}
\author[add1,add2]{Daniel Peralta}
\ead{daniel.peralta@irc.vib-ugent.be}
\author[add1]{Chris Cornelis}
\ead{chris.cornelis@ugent.be}
\address[add1]{Department of Applied Mathematics, Computer Science and Statistics, Ghent University}
\address[add2]{Data Mining and Modelling for Biomedicine group, VIB Center for Inflammation Research, Ghent University}

\address{}

\begin{abstract}
One-class classification is a challenging subfield of machine learning in which so-called data descriptors are used to predict membership of a class based solely on positive examples of that class, and no counter-examples. A number of data descriptors that have been shown to perform well in previous studies of one-class classification, like the Support Vector Machine (SVM), require setting one or more hyperparameters. There has been no systematic attempt to date to determine optimal default values for these hyperparameters, which limits their ease of use, especially in comparison with hyperparameter-free proposals like the Isolation Forest (IF). We address this issue by determining optimal default hyperparameter values across a collection of 246 one-class classification problems derived from 50 different real-world datasets. In addition, we propose a new data descriptor, Average Localised Proximity (ALP) to address certain issues with existing approaches based on nearest neighbour distances. Finally, we evaluate classification performance using a leave-one-dataset-out procedure, and find strong evidence that ALP outperforms IF and a number of other data descriptors, as well as weak evidence that it outperforms SVM, making ALP a good default choice.
\end{abstract}

\begin{keyword}
Data descriptors \sep
Hyperparameters \sep
Localised distance \sep
Nearest neighbours \sep
Novelty detection \sep
One-class classification \sep
OWA operators \sep
Semi-supervised outlier detection

\end{keyword}

\end{frontmatter}

\thispagestyle{firstpagestyle}

\section{Introduction}
One-class classification \cite{tax01oneclass}, also known as novelty detection or semi-supervised outlier detection, is an asymmetric type of binary classification between a \e{target} class and the \e{other} class. One-class classifiers, known as \e{data descriptors}, are formally distinguished from ordinary binary classifiers in that they are only trained on representatives from the target class. It is this restriction that makes one-class classification a challenging problem. In recent years, one-class classification has been applied to problems like writer identification \cite{hadjadji18two}, Twitter bot detection \cite{rodriguezruiz20oneclass} and face presentation attack detection \cite{fatemifar21clientspecific}.

One-class classification is closely related, but subtly different from unsupervised outlier detection, in which the training data is an unlabelled mixture of target and outlier instances \cite{domingues18comparative}. Data descriptors originally defined for unsupervised outlier detection are frequently repurposed for one-class classification (and vice versa), but their performance should be evaluated separately in each setting.

The evaluation of one-class classification performance of data descriptors has in the past led to partially contradictory outcomes. Janssens et al.~(2009) \cite{janssens09outlier} compared five descriptors, and found that the Support Vector Machine (SVM), Local Outlier Factor (LOF) and Localised Nearest Neighbour Distance (LNND) significantly outperform the Parzen Window (PW) and Local Correlation Integral (LOCI), as well as weak evidence that SVM and LOF outperform LNND. Swersky et al.~(2016) \cite{swersky16evaluation} replicated this test, and found instead that LNND performs no better than PW and LOCI, plus weak evidence that SVM outperforms LOF. In addition, Swersky et al.~tested 6 more descriptors, of which two of the simplest, the Nearest Neighbour Distance (NND) and Mahalanobis Distance (MD), performed among the best.

A common property of the best-performing data descriptors in these comparisons, with the exception of MD, is that they require setting one or two `magic' hyperparameters by the user, which typically control a trade-off between variance and bias. In \cite{janssens09outlier} and \cite{swersky16evaluation}, these hyperparameters are optimised for each one-class classification task through cross-validation on the training set. While this approach can be expected to result in the best possible result for that particular task, it does not fully answer the challenge raised by the authors of the Schölkopf variant of SVM to ``turn the algorithm into an easy-to-use black-box method for practitioners'' \cite{scholkopf99estimating}. For that, we need a set of sensible default values for these hyperparameters, in particular for applications where no substantial amount of training material from the other class is available for tuning. 

In contrast, a number of more recently proposed data descriptors do not require users to specify any hyperparameters. Strictly speaking, these data descriptors do have hyperparameters, but the authors have established default values that can be taken for granted. One of these is the Isolation Forest (IF) \cite{liu08isolation}, which also differs from other data descriptors because it doesn't directly model similarity to the target data, but instead scores instances heuristically by trying to isolate them. In a small-scale evaluation of unsupervised outlier detection, IF was shown to achieve better results than a number of other approaches, including LOF with a (likely sub-optimal) fixed hyperparameter choice of $k = 10$. Very recently, a number of conceptual concerns with IF have been addressed by the Extended Isolation Forest (EIF) proposal \cite{hariri21extended}, which achieved better unsupervised outlier detection results in another small-scale experiment. However, it remains unclear how well IF and EIF perform one-class classification.

Finally, a different approach is taken by Cao et al.~(2019) \cite{cao19learning}, who propose a Shrink Autoencoder (SAE) to transform the feature space in such a way that one-class classification becomes largely insensitive to the choice of hyperparameters, and even to the choice of data descriptor. We do not explore this approach in the present article. However, when paired with Centroid Distance, we can consider SAE as a data descriptor in its own right. Cao et al.~report that this outperforms LOF and SVM, especially with sparse datasets, but in this instance too the choice of hyperparameters for LOF and SVM may not have been optimal, and the selection of datasets was limited.

In the present article, we introduce a new data descriptor, Average Localised Proximity (ALP), which aims to improve upon LNND and LOF in particular. Furthermore, we address the state of affairs sketched above in two steps. Firstly, we determine optimal default hyperparameter values for NND, LNND, LOF, SVM and ALP, which allows them to be used as ``black-box methods'' like MD, IF, EIF and SAE. And secondly, we compare the performance of these nine data descriptors across 246 one-class classification tasks derived from 50 datasets.

We proceed by providing formal definitions of one-class classification in general (Section \ref{sec_occ}) and the eight existing data descriptors covered in this article in particular (Section \ref{sec_data_descriptors}). Next, we present Average Localised Proximity (Section \ref{sec_alp}) and describe the experiments that we perform (Section \ref{sec_setup}). We end with an analysis of the results (Section \ref{sec_results}) and our general conclusion (Section \ref{sec_conclusion}).

\section{One-class classification}
\label{sec_occ}

Formally, we may view one-class classification as a generalisation from a finite set of instances to a function from the entire attribute space to the unit interval (Definition \ref{one-class}).

\begin{definition}
\label{one-class}
A dataset is a pair $(A, X)$, where $A$ (the attribute space) is a set, and $X$ a finite multisubset of $A$. A data descriptor is a pair $(P, D)$, with $P$ a (possibly trivial) hyperparameter space, and $D$ a function that takes a dataset $(A, X)$ and a combination of hyperparameter values in $P$, and returns a function $A \lra [0, 1]$, which is called the data description by $D$ of $(A, X)$.
\end{definition}

For the purpose of this article, we assume that $A = \mb{R}^m$ is a linear combination of $m$ real-valued attributes.

In practical applications, $X$ is a sample drawn from some statistical population, the \e{target class}, and the goal of one-class classification is to predict, on the basis of their attribute values, whether new instances originated from this target class. For this purpose, the values in $[0, 1]$ assigned by a data description can be interpreted as confidence scores. Accordingly, a data description should ideally assign high scores to the instances in $X$, although this is not a strict requirement.

In the literature, data descriptions are often more naturally formulated as functions to some larger subspace $S \sbeq \mb{R}$, which sometimes correspond inversely with confidence. Such functions can easily be adapted to fit Definition \ref{one-class} through composition with a monotonic or anti-monotonic map $S \lra [0, 1]$. In particular, we will transform distance functions $\mb{R}^m \lra [0, \infty)$ with the map (\ref{eq_d_to_p}).

\begin{align}
z \lrma \frac{1}{1 + z} \label{eq_d_to_p}
\end{align}

It may be desirable to make a more definite statement by using a threshold $\alpha$, and predicting that all instances with a score greater than or equal to $\alpha$ belong to the target class. The choice of $\alpha$ determines a trade-off between false positive and false negative predictions, and should therefore be informed by the context in which the data description is deployed.

We say that two data descriptions are equivalent if they can be transformed into each other through a strictly order-preserving map on $[0, 1]$. Such a map is also a map between thresholds. 

We can evaluate the performance of a data description using a test set $Y \sbeq A$ containing both elements drawn from the target class (but not contained in $X$), as well as other instances. A common evaluation metric is classification accuracy, but this is dependent on a choice of threshold. The Area Under the Receiver Operating Characteristic (AUROC) \cite{hanley82meaning,bradley97use} is a better evaluation metric of the data description as such, since it encodes the ability of the data description to separate target instances from other instances. It directly corresponds to the chance that a random target instance receives a higher score than a random other instance. In practice, several distinct test sets are often employed as part of a cross-validation scheme, in which case the performance of the data description may be summarised with the mean AUROC.

One-class classification as described above is also sometimes called novelty detection or semi-supervised outlier detection. There is a closely related machine learning problem known as unsupervised outlier or anomaly detection, where the assumption is that most but not all of the instances in $X$ are drawn from the target class, and the task is not to generalise $X$ to $A$, but to identify the elements in $X$ that do not belong to the target class. The distinction between novelty and unsupervised outlier detection may not always be clear in practice, since on the one hand many unsupervised outlier detectors can also be used to assign scores to instances outside of $X$, and on the other hand novelty detectors may also assign low scores to some atypical instances within $X$. The most important functional distinction is whether $X$ may contain instances not drawn from the target class.

In this article, we are concerned with one-class classification as novelty detection. Nevertheless, we will discuss some data descriptors that were originally proposed for unsupervised outlier detection, but have since been repurposed.

\section{Existing data descriptors}
\label{sec_data_descriptors}
We now discuss a number of data descriptors that have been evaluated favourably by Janssens et al.~\cite{janssens09outlier} and/or Swersky et al.~\cite{swersky16evaluation}, as discussed in the Introduction, as well as the recent Isolation Forest and Extended Isolation Forest proposals. Throughout, we assume a dataset $(\mb{R}^m, X)$ and a generic instance $y \in \mb{R}^m$, and define a descriptor through the action of its description of $X$ on $y$. For the descriptors based on nearest neighbour distances, we use $\NN_k(y)$ to denote the $k$th nearest neighbour of $y$ in $X$, excluding $y$ itself if $y$ is explicitly drawn from $X$. We will define the respective hyperparameter spaces informally, as a choice of hyperparameters.

\subsection{Nearest Neighbour Distance}

Nearest Neighbour Distance (NND) is conceptually very simple, and goes back to at least \cite{knorr97unified}. In its general form, it requires a choice of a dissimilarity measure $d$ and a positive integer $k$. It is based on the score $d_k(y) = d(y, \NN_k(y))$, from which we obtain a data description through composition with (\ref{eq_d_to_p}). While this is not scale-invariant, different scalings lead to equivalent descriptions.

\subsection{Localised Nearest Neighbour Distance}
The argument for Localised Nearest Neighbour Distance (LNND) \cite{ridderde98experimental,tax98outlier} is that distance in the attribute space should not be valued equally everywhere, but that it should instead be compared to the local distance between nearby training instances. Thus, for a choice of dissimilarity $d$ and a positive integer $k$, we can define the local distance $d^2_k(y) = d_k(\NN_k(y))$ relative to a point $y \in \mb{R}^m$ and the localised distance $ld_k(y) = d_k(y)/d^2_k(y)$, which we compose with (\ref{eq_d_to_p}) to obtain a data description.

\subsection{Local Outlier Factor}
Local Outlier Factor (LOF), originally proposed for unsupervised outlier detection \cite{breunig00lof}, is also based on nearest neighbour distances, and also involves a form of localisation. For a choice of dissimilarity $d$ and positive integer $k$, it derives from $d$ the (non-symmetric) \e{reachability} distance $rd_k$ (\ref{eq_rd}). The goal is to ``significantly reduce'' the ``statistical fluctuations'' that occur when query instances lie very close to target instances. It achieves this by effectively cancelling out all distances that are smaller than a certain threshold, determined by local nearest neighbour distances.

\begin{align}
rd_k(y, x) &= \max(d(y, x), d_k(x)), \label{eq_rd}
\end{align}

LOF then aggregates and inverts a range of reachability distance values to obtain the local reachability density $\lrd_k$ (\ref{eq_lrd}), and localises and aggregates a range of local reachability density values to end up with the local outlier factor $\lof_k$ (\ref{eq_lof}), which we compose with (\ref{eq_d_to_p}) to obtain a data description.

\begin{align}
\lrd_k(y) &= \frac{1}{\frac{1}{k}\sum_{i \leq k} rd_k(y, \NN_i(y))}, \label{eq_lrd}\\
\lof_k(y) &= \frac{1}{k}\sum_{j \leq k} \frac{\lrd_k(\NN_j(y))}{\lrd_k(y)}. \label{eq_lof}
\end{align}

\subsection{Mahalanobis Distance}
One of the oldest approaches to novelty detection is to assume that the target class samples are drawn from a multivariate Gaussian Distribution. The atypicality of an instance under this distribution is given by the Mahalanobis Distance (MD) \cite{mahalanobis36generalized} to this distribution (\ref{eq_mahalanobis}), where $\mu$ and $S$ are the mean and the covariance matrix of the training instances.

\begin{align}
D(y) = \sqrt{(y - \mu)^T S^{-1} (y - \mu)}, \label{eq_mahalanobis}
\end{align}

Mahalanobis distance generalises distance from the mean in terms of standard deviations in a univariate Gaussian distribution. Squared Mahalanobis distance follows a $\chi^2$-distribution with $m$ degrees of freedom, and we obtain a $p$-value (\ref{eq_dd_md}) by applying its cumulative distribution function $F$ and subtracting from 1.

\begin{align}
p(y) = 1 - F(D(y)^2). \label{eq_dd_md}
\end{align}

This $p$-value is a natural choice for a data description, but it approaches 0 very quickly, making it computationally difficult to distinguish between $p$-values of large Mahalanobis distances (they are all rounded to 0). Therefore, in order not to unduly limit the discriminative power of MD, we instead compose $D$ with (\ref{eq_d_to_p}) in this article.

\subsection{Support Vector Machine}

There exist two, practically equivalent, adaptations of the soft-margin Support Vector Machine (SVM) \cite{cortes95support} to one-class classification. Both allow the use of a kernel $k: \mb{R}^m \times \mb{R}^m \lra \mb{R}$ to transform the feature space in which we obtain the solution via an implicit map $\phi: \mb{R}^m \lra Z$ to some inner product space $Z$.

The Tax variant \cite{tax99data,tax99support} fits a hypersphere of minimal volume around the training instances by solving the optimisation problem (\ref{eq_svm_t_primal}), with dual (\ref{eq_svm_t_dual}), for a choice of $C \in [0, \infty)$ and a choice of kernel $k$. The instances $x_i$ with corresponding non-zero values of $\alpha_i$ are the support vectors --- these span a hypersphere centred at some point $a \in Z$ with radius $R^2$. The parameter $C$ determines how many training instances remain outside the hypersphere. The decision function is the distance $d_T$ to $a$ (\ref{eq_svm_t_distance}).

\begin{align}
\min_{a \in Z, R \in \mb{R}, \xi \in \mb{R}_{\geq 0}^n} R^2 + C \sum_i \xi_i, \text{\quad\quad with } \forall i: \left\lVert \phi(x_i) - a\right\rVert^2 \leq R^2 + \xi_i \label{eq_svm_t_primal}\\
\min_{\alpha \in [0, C]^n} \sum_i \alpha_i k(x_i, x_i) - \sum_{i, j} \alpha_i\alpha_j k(x_i, x_j), \text{\quad\quad with } \sum_i \alpha_i = 1 \label{eq_svm_t_dual}\\
d_T(y) = k(y, y) - 2\sum_i \alpha_i k(y, x_i) + \sum_{i, j} \alpha_i\alpha_j k(x_i, x_j) \label{eq_svm_t_distance}
\end{align}

The Schölkopf variant \cite{scholkopf99estimating,scholkopf01estimating} fits a hyperplane to separate the training instances from the origin, at a maximum distance from the origin, by solving the optimisation problem (\ref{eq_svm_s_primal}), with dual (\ref{eq_svm_s_dual}), for a choice of $\nu \in (0, 1]$ and a choice of kernel $k$.  The instances $x_i$ with corresponding non-zero values of $\alpha_i$ are the support vectors --- these span a hyperplane with distance $\rho$ to the origin. The parameter $\nu$ determines how many training instances remain on the wrong side of the hyperplane. The decision function is the signed distance $d_S$ to this hyperplane (\ref{eq_svm_t_distance}), with negative values on the side of the origin.

\begin{align}
\min_{w \in Z, \rho \in \mb{R}, \xi \in \mb{R}_{\geq 0}^n} \frac{1}{2} \left\lVert w \right\rVert^2 + \frac{1}{\nu n} \sum_i \xi_i - \rho, \text{\quad\quad with } \forall i: w \cdot \phi(x_i) \geq \rho - \xi_i \label{eq_svm_s_primal}\\
\min_{\alpha \in [0, \frac{1}{\nu n}]^n} \frac{1}{2}\sum_{i, j} \alpha_i\alpha_j k(x_i, x_j), \text{\quad\quad with } \sum_i \alpha_i = 1 \label{eq_svm_s_dual}\\
d_S(y) = \sum_i \alpha_i k(x_i, y) - \rho. \label{eq_svm_s_distance}
\end{align}

Both SVM variants have been shown to produce the best overall results with the Gaussian kernel (\ref{eq_gaussian_kernel}), which requires a choice for the \e{kernel width} $c$.

\begin{align}
k(x, y) = e^{-\frac{\left\lVert x - y \right\rVert^2}{c}} \label{eq_gaussian_kernel}
\end{align}

For kernels $k$ for which $k(y, y)$ is constant, like the Gaussian kernel, the respective optimisation problems become equivalent (with reparametrisation $C = \frac{1}{\nu n}$), resulting in the same set of support vectors \cite{scholkopf01estimating,tax04support}, and the decision functions $d_T$ and $d_S$ are monotonic linear transformations of each other, leading to equivalent data descriptions. In this article we use a data description based on the Schölkopf variant (\ref{eq_dd_svm}), for which we have access to an implementation. Instances on the hyperplane receive a score of $0.5$.

\begin{align}
y \lrma \frac{1}{2}\left(\frac{d_S(y)}{\left\lvert d_S(y) \right\rvert + 1} + 1\right). \label{eq_dd_svm}
\end{align}

With the Gaussian kernel, the Schölkopf variant of SVM has two hyperparameters that need to be tuned, $\nu$ and $c$.

\subsection{Isolation Forest}
Isolation Forest (IF) \cite{liu08isolation} is an adaptation of the Random Forest classifier for one-class classification. Its central idea is that instances that are more isolated from the target class should be easier to separate from the training instances. This idea is modelled by constructing randomised search trees on the target data, and measuring the average number of steps required to pass through these trees.

For positive integers $t$ and $\psi$, IF creates $t$ binary incomplete search trees of height at most $\left\lceil\log_2(\psi)\right\rceil$ using $t$ random subsamples of $\psi$ instances of $X$, in which each node splits the remaining instances by randomly selecting an attribute and a corresponding value within the range of remaining values, until the maximum tree height is reached.

The expected average path length in a tree of $i$ instances is expressed by $c$ (\ref{eq_if_c}), where $H_{i - 1}$ is the $i-1$th harmonic number.

\begin{align}
c(i) = 2H_{i - 1} - 2(i - 1)/i \label{eq_if_c}
\end{align}
 
For each tree $T$, define $h_T(y)$ as the sum of the path length of $y$ in $T$ and $c(j)$, where $j$ is the subsample size remaining in the final node of $y$ in $T$. The rationale behind limiting tree height and estimating the remaining path length is that this limits the number of search steps, while it mostly affects target class instances.

From this we obtain an anomaly score $s$ in $[0, 1]$ (\ref{eq_if_anomaly}), which we subtract from 1 to obtain a data description.

\begin{align}
s(y) = 2^{-\frac{\frac{1}{t}\sum_{T}h_T(y)}{c(\psi)}}, \label{eq_if_anomaly}
\end{align}

The performance of IF initially increases with the number of trees $t$ and the subsample size $\psi$, but eventually converges, and \cite{liu08isolation} experimentally finds that these hyperparameters can safely be set to 100 and $\min(256, n)$ respectively. Hence, a significant advantage of IF is that it has no tuning parameters.

\subsection{Extended Isolation Forest}
Extended Isolation Forest (EIF) \cite{hariri21extended} is a slight modification of IF, which only changes the construction of the trees. It is motivated by the observation that IF seems to produce counter-intuitive results with simple distributions. In particular, IF assigns much lower anomaly scores to instances that are far removed from the training set along a single feature dimension (but not the others) than to instances that are slightly removed from the training set in several feature dimensions. This is due to the fact that the splits in the binary trees correspond to hyperplanes in the attribute space that always lie parallel to $m - 1$ feature dimensions.

EIF solves this bias by splitting its trees along hyperplanes with a randomly chosen slope and an intersect randomly drawn from the range determined by the remaining instances.

\subsection{Shrink Autoencoder}
The problem of one-class classification can also be approached with autoencoders. These neural networks learn a latent representation of a dataset by forcing the data through a bottleneck layer with fewer features than the input. After this encoding step, the data is decoded again by applying the same weights in reverse, and the whole model is trained by minimising the resulting reconstruction error. After training is completed, the reconstruction error of new instances can be used to define a data descriptor. However, Cao et al.~(2019) \cite{cao19learning} argue that better results can be obtained by working directly with the latent representation of an autoencoder, if we force this representation to conform to a compact distribution. For this purpose, Cao et al.~propose the Shrink Autoencoder (SAE) and the Dirac Delta Variational Autoencoder (DVAE). We only consider SAE in this paper, as its reported results are slightly better and it has proven easier to implement.

SAE is an autoencoder with five hidden layers, in which the number of features is linearly reduced from $m$ to $\left\lfloor\sqrt{m}\right\rceil + 1$ --- the number of latent features of the central layer --- and increased back to $m$. In addition to the reconstruction error, the loss function also incorporates $L_2$ regularisation on the central layer, thereby directing the learning process towards latent representations that are distributed closely around the origin.

By substituting the latent for the original representation of the data, SAE can be used as a preprocessing step in combination with any other data descriptor. Cao et al.~show that this is beneficial, especially for sparse datasets. Moreover, due to the regular shape of the latent representation, the choice of hyperparameters and even the choice of data descriptor becomes largely irrelevant. Therefore, Cao et al.~propose that one may use SAE in combination with a simple centroid data descriptor, which measures the distance to the mean after rescaling by the respective standard deviations in each dimension. It is this combination that we will test in the present article.

\section{Average Localised Proximity}
\label{sec_alp}

One remarkable finding of the empirical comparison in \cite{swersky16evaluation} is that NND outperforms LNND. Yet the principle behind LNND seems sensible: if a dataset is more densely distributed in one part of the attribute space than in another, we should adjust our expectations regarding nearest neighbour distance accordingly. So we may ask what causes LNND to perform badly.

The most obvious problem with LNND is that its measure for local nearest neighbour distance is not very robust, since it is determined by a single distance in the training set. In every natural dataset, there is random variation in the distances between instances, and this directly translates into random variation of the LNND scores of test instances. Secondly, while it seems elegant to localise against the $k$th nearest neighbour distance of the $k$th nearest neighbour of a test instance, it is not clear why the $k$th nearest neighbour distance of the closest training instance isn't more relevant. And finally, LNND only considers neighbour distances for a single value of $k$, whereas it would be more robust to aggregate over different values for $k$.

These problems are addressed by LOF, but LOF has issues of its own. The conceptual motivation for its slightly convoluted amalgam of localisation and reachability is not entirely clear. LOF seems to take localisation one step too far, with three rounds of averaging that require calculating the distance to the $k$th neighbour of the $i$th neighbour of the $j$th neighbour of a test instance $y$. There is also a degree of arbitrariness to its application of reachability, as it uses $k$th neighbour distances as a threshold for $i$th neighbour distances, with $k$ generally larger than $i$. More fundamentally, it is not clear that reachability is a good idea, since it may discard useful information by cancelling out small distances.

As an alternative, we propose Average Localised Proximity (ALP), a new data descriptor that is more robust than LNND, yet conceptually simpler than LOF (Definition \ref{def_alp}). It uses an Ordered Weighted Averaging (OWA) operator \cite{yager88ordered} (Definition \ref{def_owa}) to obtain a soft maximum.

\begin{definition}
\label{def_owa}
Let $w$ be a weight vector of length $k$, with monotonically decreasing values in $[0, 1]$ that sum to 1. The \e{Ordered Weighted Averaging} operator $\owa{w}$ induced by $w$ transforms a collection $Y = \left\{y_i\right\}_{i \leq k}$ of values in $\mb{R}$ into the weighted sum $\owa{w}_{i \leq k} y_i = \sum_{i \leq k} w_i\cdot y_{o(i)}$, where $o(i)$ is the index value of the $i$th largest element in $Y$.
\end{definition}

\begin{definition}
\label{def_alp}
Let $(\mb{R}^m, X)$ be a dataset, let $d$ be a choice of dissimilarity function on $\mb{R}^m$, let $k, l \in \mb{N}$ be choices of positive integers, and let $w^k, w^l$ be choices of weight vectors of length $k$ and $l$ respectively. For each $i \leq k$, define $D_i(y)$ (\ref{eq_alp_ld}), the local $i$th neighbour distance relative to $y$, and from this, $\lp_i(y)$ (\ref{eq_alp_lp}), the localised proximity of $y$. Then the \e{average localised proximity} $\alp(y)$ of $y$ is the ordered weighted average of these values (\ref{eq_alp}).

\begin{align}
D_i(y) &= \sum_{j \leq l}w^l_j \cdot d_i(\NN_j(y)). \label{eq_alp_ld}\\
\lp_i(y) &= \frac{D_i(y)}{D_i(y) + d_i(y)}. \label{eq_alp_lp}\\
\alp(y) &= \owa{w^k}_{i \leq k} \lp_i(y). \label{eq_alp}
\end{align}
\end{definition}

\begin{figure}
\tikzset{
block/.style={
    draw, ultra thin
},
line/.style={>=latex,->, shorten <=0.5em, shorten >=0.5em}
}
\begin{tikzpicture}
\matrix (m)[matrix of nodes, column  sep=0cm,row  sep=2cm, align=center, nodes={anchor=center} ]{
    |[block, text width=0.5em, label=Query instance]| {\scriptsize y} &
    |[block, text width=3em]|  {\scriptsize$\NN_1(y)$\\$\NN_2(y)$\\$\vdots$\\$\NN_l(y)$} \\
    |[block, text width=9em]| {\scriptsize $d_1(y)$, $d_2(y)$, $\dots$, $d_k(y)$} &
    |[block, text width=16em]| {\scriptsize
    $d_1(\NN_1(y))$, $d_2(\NN_1(y))$, $\dots$, $d_k(\NN_1(y))$\\
    $d_1(\NN_2(y))$, $d_2(\NN_2(y))$, $\dots$, $d_k(\NN_2(y))$\\
    $\vdots$\\
    $d_1(\NN_l(y))$, $d_2(\NN_l(y))$, $\dots$, $d_k(\NN_l(y))$}          &                                             \\
    |[block, text width=10em, label=Localised proximities]| {\scriptsize$\lp_1(y)$, $\lp_2(y)$, $\dots$, $\lp_k(y)$} &
    |[block, text width=10em, label=Local distances]| {\scriptsize$D_1(y)$, $D_2(y)$, $\dots$, $D_k(y)$} \\
    |[block, text width=2em, label=Average localised proximity]| {\scriptsize$\alp(y)$} \\
};
\draw [line] (m-1-1) -- node[midway,below]{\scriptsize $l$ nearest neighbours} (m-1-2);
\draw [line] (m-1-1) -- node[midway,fill=white]{\scriptsize $k$ nearest neighbour distances} (m-2-1);
\draw [line] (m-1-2) -- node[midway,fill=white]{\scriptsize $k$ nearest neighbour distances} (m-2-2);
\draw [line, shorten >=1.5em] (m-2-1) -- node[midway]{} (m-3-1);
\draw [line, shorten >=1.5em] (m-2-2) -- node[midway,fill=white]{\scriptsize Weighted average with $w^l$} (m-3-2);
\draw [line, shorten >=1.5em] (m-3-1) -- node[midway,fill=white]{\scriptsize Ordered weighted average with $w^k$} (m-4-1);
\draw [line] (m-3-2) -- node[midway,right]{} (m-3-1);
\end{tikzpicture}
\caption{Schematic illustration of the calculation of the average localised proximity of a query instance $y$.}
\label{fig_alp_diagram}
\end{figure}
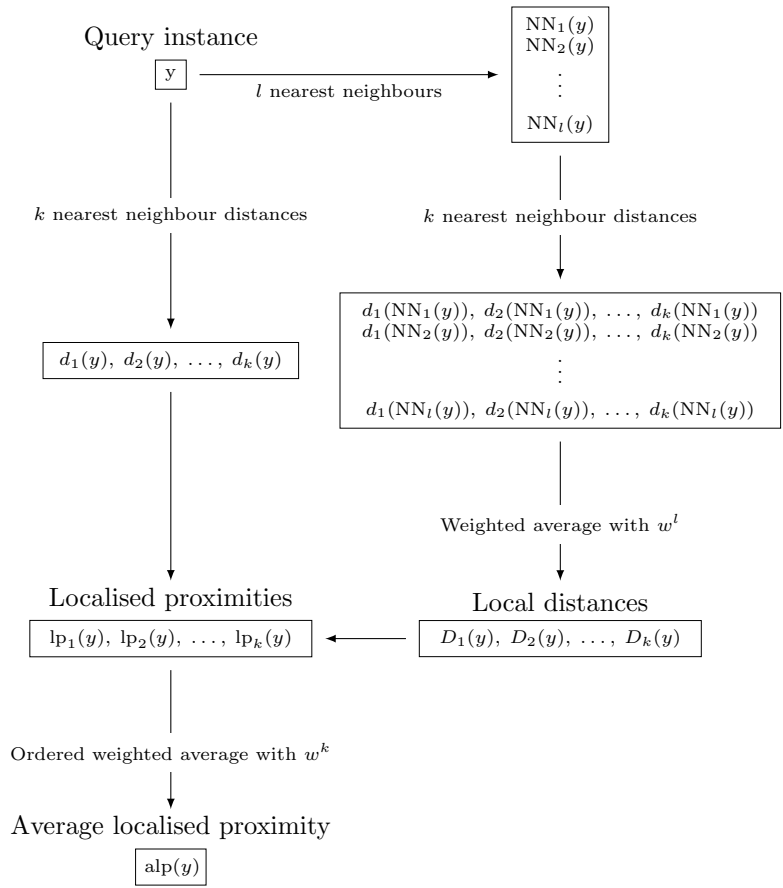

As illustrated in Figure \ref{fig_alp_diagram}, ALP features aggregation on two levels. Local nearest neighbour distance is determined with a weighted sum over a section of the training set. By choosing monotonically decreasing weights, we let the contribution of training instances to this weighted sum decrease with distance to $y$. The choice of weight vector determines the amount of localisation, which can be seen as a trade-off between variance and bias. For $k = 1$, we recover LNND if also $l = 1$, whereas if $l = n$ and all weights equal $\frac{1}{n}$, all distances are localised against the training set mean and we obtain a data description that is equivalent to NND.

We thus obtain $k$ localised distance values, which we can interpret as representing different scales. Inspired by fuzzy rough set theory \cite{cornelis10ordered}, we first transform these into proximity values in $[0, 1]$, and then apply an OWA operator to obtain a soft maximum. With suitable weights, the OWA operator offers a combination of flexibility and robustness, emphasising the scales at which test instances have the greatest proximity to the target class, without being completely determined by any single scale. The impact of scales with small proximities is also reduced by our choice to aggregate after transforming values from $[0, \infty]$ to $[0, 1]$, rather than vice-versa.

The choice of weight vector offers further opportunity for optimisation. However, our previous experience in fuzzy rough set theory \cite{lenz20scalable} has been that the effect of this choice may in fact be limited, and that linearly decreasing weights constitute a good default. These are defined, for $p = k, l$, as $\frac{p}{p(p + 1)/2}, \frac{p - 1}{p(p + 1)/2}, \dots, \frac{1}{p(p + 1)/2}$ (the denominator is the $p$th triangular number). Similarly, a good enough default choice for $d$ is the Manhattan metric. Therefore, the only other hyperparameters to be specified by the user are $k$ and $l$.

\begin{figure}
\centering
\includegraphics[width=\linewidth]{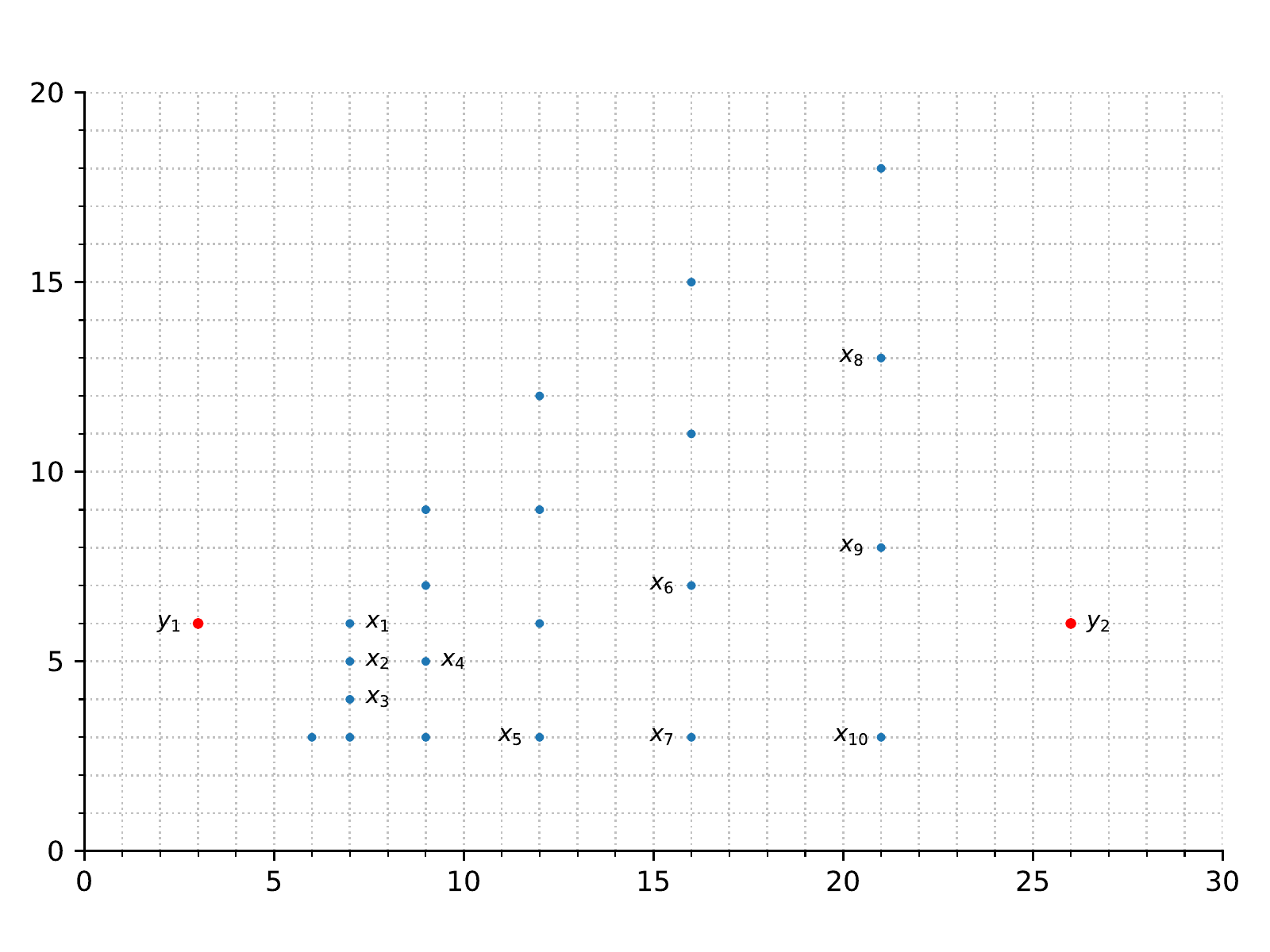}%
\caption{One-class classification example, with training instances ($x_1, x_2, \dots$) and query instances ($y_1$ and $y_2$).}
\label{fig_alp_example}
\end{figure}

\begin{table}
\centering
\caption{Selected values used in the application of ALP to the one-class classification example in Figure \ref{fig_alp_example}.}
\label{tab_alp_example_vals}
\subfloat[]{\begin{tabular}{cccccccc}
\toprule
 & & $\NN_1$ & $\NN_2$ & $\NN_3$ & $d_1$ & $d_2$ & $d_3$\\
\midrule
$y_1$ & & $x_1$ & $x_2$ & $x_3$ & 4 & 5 & 6\\
$y_2$ & & $x_9$ & $x_{10}$ & $x_6$ & 7 & 8 & 11\\
$x_1$ & & $x_2$ & $x_3$ & $x_4$ & 1 & 2 & 3\\
$x_2$ & & $x_1$ & $x_3$ & $x_4$ & 1 & 1 & 2\\
$x_9$ & & $x_8$ & $x_{10}$ & $x_6$ & 5 & 5 & 6\\
$x_{10}$ & & $x_7$ & $x_9$ & $x_5$ & 5 & 5 & 9\\
\bottomrule
\end{tabular}
\label{tab_alp_example_1}}\\
\subfloat[]{\begin{tabular}{cccccccc>{\bfseries}c}
\toprule
 & & $D_1$ & $D_2$ & $D_3$ & $\lp_1$ & $\lp_2$ & $\lp_3$ & $\alp$ \\
\midrule
$y_1$ & & 1.00 & 1.67 & 2.67 & 0.20 & 0.25 & 0.31 & 0.27\\
$y_2$ & & 5.00 & 5.00 & 7.00 & 0.42 & 0.38 & 0.39 & 0.40\\
\bottomrule
\end{tabular}
\label{tab_alp_example_2}}

\end{table}

We illustrate the application of ALP with a toy example (Figure \ref{fig_alp_example}), with two instances to be scored ($y_1$ and $y_2$) and a number of training instances ($x_1, x_2, \dots$). We choose $k = 3$ and $l = 2$. Accordingly, we obtain $w^k = \left\langle \frac{3}{6}, \frac{2}{6}, \frac{1}{6}\right\rangle$ and $w^l = \left\langle \frac{2}{3}, \frac{1}{3}\right\rangle$.

A relevant selection of nearest neighbours and nearest neighbour distances is listed in Table \ref{tab_alp_example_1}, while Table \ref{tab_alp_example_2} contains the resulting local distances relative to $y_1$ and $y_2$, their localised proximities and their average localised proximity scores. The local distances are calculated as the weighted sum of the nearest neighbour distances of $x_1$ and $x_2$ for $y_1$, and $x_9$ and $x_{10}$ for $y_2$, weighted with $\frac{2}{3}$ and $\frac{1}{3}$ respectively. The final scores are obtained by sorting the localised proximities of $y_1$ and $y_2$ and from large to small, and taking their sum with weights $\frac{3}{6}$, $\frac{2}{6}$ and $\frac{1}{6}$.

\section{Experimental setup}
\label{sec_setup}
A large part of the one-class classification problems used by the two previous comparison studies \cite{janssens09outlier,swersky16evaluation} mentioned in the Introduction were created by David Tax\footnote{http://homepage.tudelft.nl/n9d04/occ/index.html} on the basis of binary or multiclass classification datasets in the UCI machine learning repository \cite{dua19uci}, by selecting one decision class as the target class, and combining the rest to form the other class. We apply the same procedure to a full selection of all datasets from the UCI machine learning repository that represent real data, consist of only numerical, ordinal or binary attributes and that are not too large to make evaluation impractical. By matching the variation contained in the UCI machine learning repository, we aim to achieve a certain degree of representativeness. In total, we have selected 50 datasets, containing 246 individual decision classes, with up to \num{194198} target class instances and up to 649 attributes (Table \ref{tab_default_auroc})\footnote{A few of the datasets have a handful of instances with missing attribute values. These instances were omitted.}. We will also consider the sparsity of the target datasets, which we define as the rate of attribute values equal to the respective mode.\footnote{Our definition differs slightly from that of Cao et al.~(2019) \cite{cao19learning}, who take the rate of zeros. The motivation for our choice is that it makes sparsity independent of the encoding of the data.}

For each descriptor, dataset and class, we perform one-class classification with 5-fold stratified cross validation. We evaluate the performance of one-class classification with the mean Area Under the Receiver Operator Characteristic (AUROC) across folds. We have chosen AUROC because it encodes how well a given data descriptor is able to separate the target and the other classes.

Some of the data descriptors in Sections \ref{sec_data_descriptors} and \ref{sec_alp} are sensitive to the relative scale of the attributes. To ensure that the potential contribution of each attribute is approximately equal, we rescale the instances within each fold, using only information from the target class instances in the training set. Since we are dealing with a large variety of target class distributions, including some that may be heavily skewed, we have chosen a robust measure of scale: the interquartile range \cite{rousseeuw93alternatives}. By rescaling attribute values to match the interquartile range of the target class, the central half of all attribute values are brought to the same scale, which is not influenced by extreme values and incidental outliers.

For EIF, we use the implementation provided by the authors,\footnote{https://github.com/sahandha/eif} while for SVM and IF, as well as for nearest neighbour searches, we use implementations provided by \verb|scikit-learn| \cite{pedregosa11scikitlearn}. We have implemented SAE in Keras and TensorFlow based on the code provided by the authors.\footnote{https://github.com/vanloicao/SAEDVAE}, and we use our own Python wrapper \verb|fuzzy-rough-learn| \cite{lenz20fuzzyroughlearn} for MD, NND, LNND, LOF and ALP.

\begin{table}
\centering
\caption{Data descriptors with hyperparameters and reparametrisations, with $n$ the size of the target class and $m$ the number of attributes. Hyperparameters $k$ and $l$ rounded to the nearest integer in the range $[1, n-1]$.}
\label{tab_reparametrisations}
\begin{tabular}{llllr}
\toprule
Data descriptor & Hyperparameter & Reparametrisation & Resolution & Window size \\
\midrule
NND  & $k$   &           & 1    & 3\\
LNND & $k$   & $a\log n$ & 0.01 & 101\\
LOF  & $k$   & $a\log n$ & 0.01 & 101\\
SVM  & $\nu$ &           & 0.1  & 11\\
     & $c$   & $c'm$     & 0.1  & $\times$ 11\\
ALP  & $k$   & $a\log n$ & 0.1  & 11\\
     & $l$   & $b\log n$ & 0.1  & $\times$ 11\\
\bottomrule
\end{tabular}
\end{table}

Of the data descriptors discussed in this article, NND, LNND, LOF, SVM and ALP have hyperparameters that need to be tuned (Table \ref{tab_reparametrisations}). We evaluate the performance of data descriptor with a specific combination of hyperparameter values by first calculating the mean AUROC for each dataset, and then the mean of these mean values across datasets. This ensures that datasets with a large number of classes do not dominate the final result.

During initial experimentation, we found that we obtain better overall results by reparametrising some hyperparameters in terms of the number of instances or features. We also found that for small hyperparameter differences, the response in average AUROC becomes noisy. We can interpret this noisiness as a natural limit on the resolution with which it makes sense to determine optimal default hyperparameter values, at least on the basis of our current selection of datasets.\footnote{In the case of SVM, another more practical limit is computation time. This is less of an issue for the other data descriptors because we can re-use nearest neighbour searches, and because the number of possible values for $k$ and $l$ is finite.} To make the results more robust, we apply a rolling mean with a centered window (two-dimensional in the case of SVM and ALP). The reparametrisations, the chosen resolution and the window size are also listed in Table \ref{tab_reparametrisations}.

In addition to the hyperparameters listed in Table \ref{tab_reparametrisations}, NND, LNND and LOF also require a choice of dissimilarity measure. We will see in Section \ref{sec_results} that NND, LNND and LOF obtain near-uniformly better results with the Manhattan than with the Euclidean metric, so we decided to simplify the rest of the experiments by limiting them to the Manhattan metric.

We will first use this setup to identify and report recommended default hyperparameter values for each data descriptor.

Next, we compare the performance of all data descriptors in Sections \ref{sec_data_descriptors} and \ref{sec_alp} as black-box classifiers with predetermined hyperparameter values. To ensure that the comparison is fair and the results generalise to other datasets, we use a leave-one-dataset-out scheme, where for each dataset we use those hyperparameter values that maximise AUROC across the other datasets. Our first question is whether our proposal, ALP, performs better than the existing data descriptors. But we will also have the opportunity to test whether any of the existing data descriptors can be said to outperform each other.

It is common in machine learning to determine whether a newly proposed classifier performs better than rival proposals by performing a Friedman test on the mean ranks \cite{demsar06statistical}. However, the appropriateness of this test has been questioned by Benavoli et al.~(2016) \cite{benavoli16should}, because the resulting p-values may be unduly inflated or deflated by the selection of classifiers. Instead, Benavoli et al.~recommend a Wilcoxon signed-rank test followed by a correction for the family-wise error.

An additional issue is that our observations contain one-class classification problems based on the same dataset. Even though data descriptors only use training data from the target class, which differs for each observation, they are still based on the same type of data, and we cannot assume that these observations are completely independent. To address this, we choose to apply a clustered Wilcoxon signed-rank test \cite{rosner06wilcoxon}, as implemented in the R package \verb|clusrank| \cite{jiang20wilcoxon}.

For each data descriptor, we test whether it performs better than the eight other data descriptors with a series of eight one-sided clustered Wilcoxon signed-rank tests. We apply the Holm-Bonferroni method \cite{holm79simple} to correct for family-wise error. The Holm-Bonferroni method indexes the eight uncorrected $p$-values from small to large, and then derives the corrected values as $\tilde{p}_i = \max_{j \leq i} (9 - j) \cdot p_j$.

Finally, we will also have a look at the ability of the data descriptors in this article to scale to large datasets, by measuring the construction and query times of one-class classification with a series of training sets of increasing size. For this purpose, we draw random samples from the large \e{higgs} dataset \cite{baldi14searching} in the UCI machine learning repository. At each size, we report the average single-threaded computation time based on five such samples. We query with a test set of 1024 instances, and report the average query time per instance.

\section{Results and analysis}
\label{sec_results}

\begin{figure}
\centering\subfloat[NND]{\includegraphics[width=\linewidth]{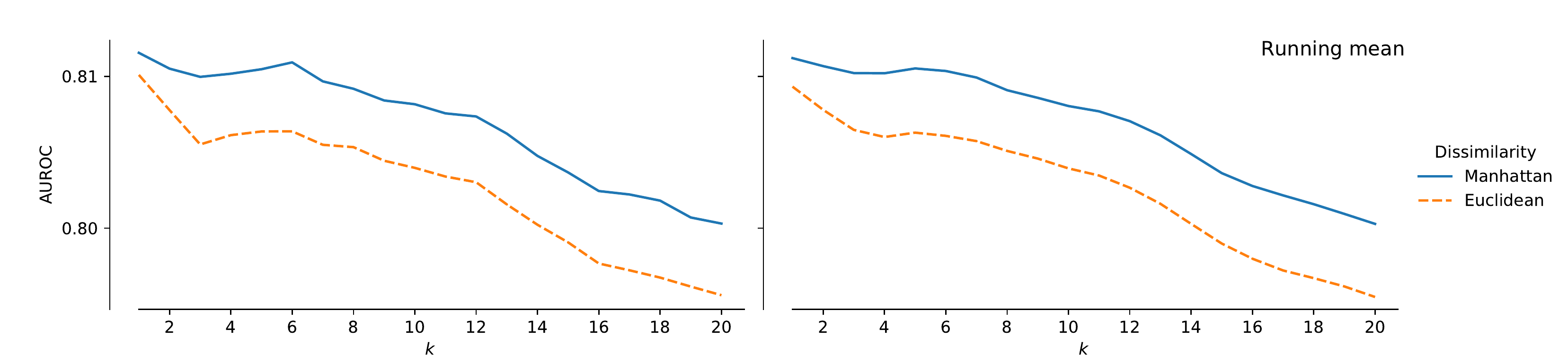}%
\label{fig_nnd_auroc}}\\
\subfloat[LNND]{\includegraphics[width=\linewidth]{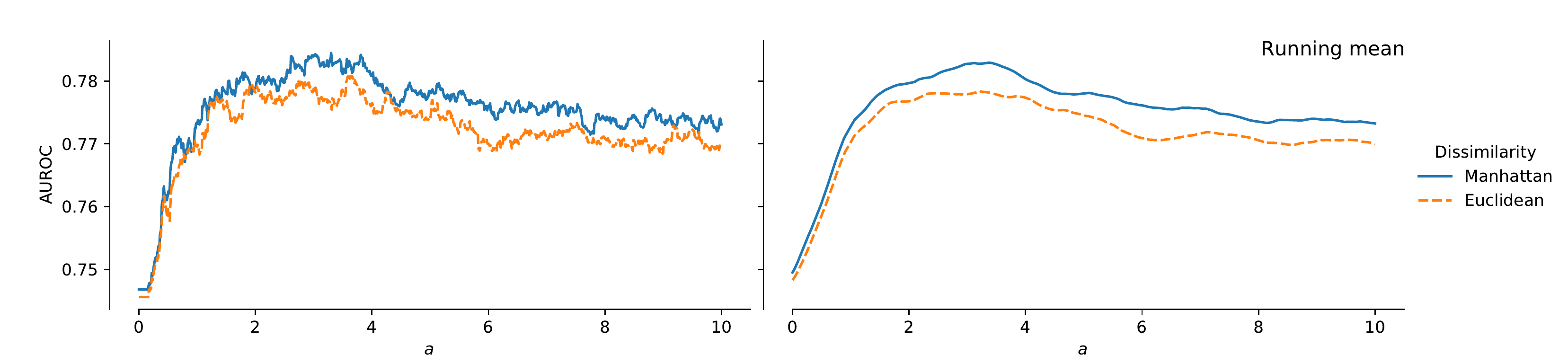}%
\label{fig_lnnd_auroc}}\\
\subfloat[LOF]{\includegraphics[width=\linewidth]{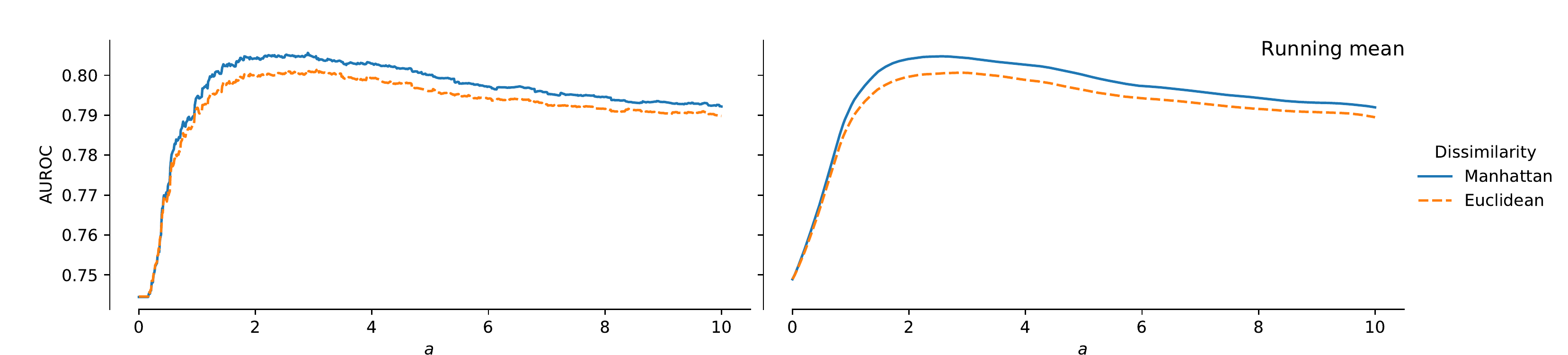}%
\label{fig_lof_auroc}}\\
\subfloat[SVM]{\includegraphics[width=\linewidth]{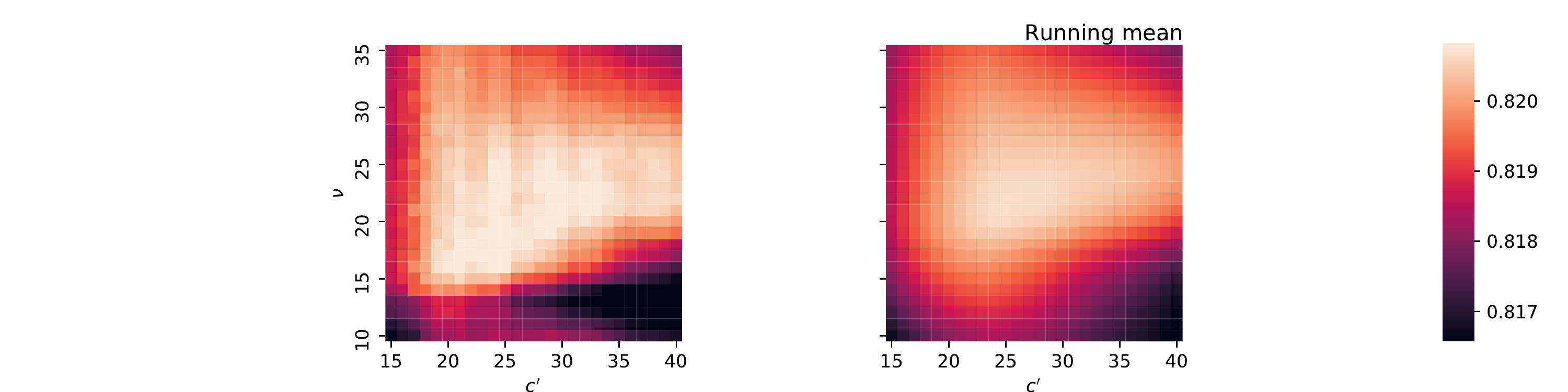}%
\label{fig_svm_auroc}}\\
\subfloat[ALP]{\includegraphics[width=\linewidth]{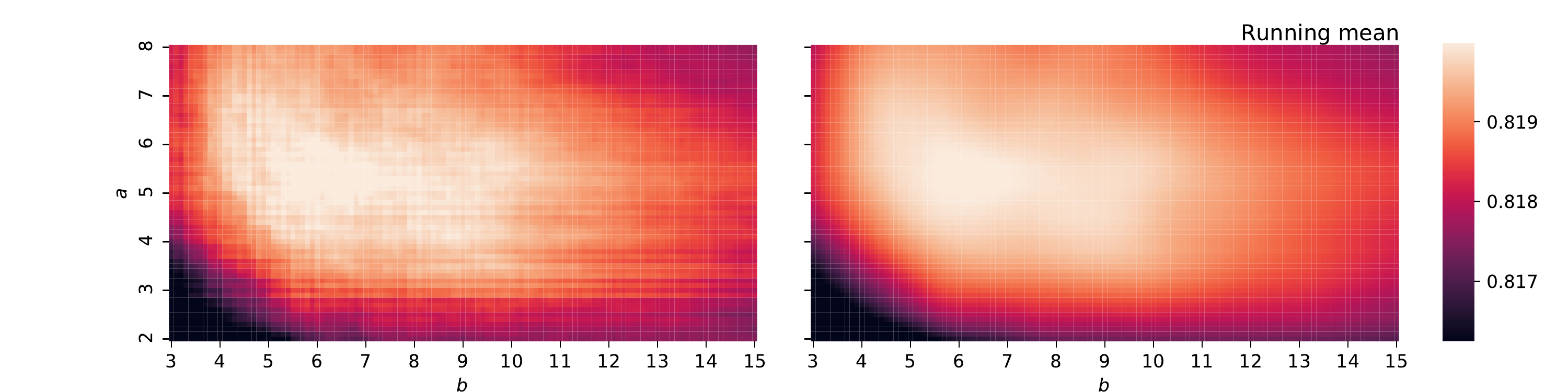}%
\label{fig_alp_auroc}}\\
\caption{Weighted mean AUROC of data descriptors across their respective hyperparameter spaces.}
\label{fig_auroc}
\end{figure}

\begin{table}
\centering
\caption{Optimal default hyperparameter values of data descriptors, with $n$ the size of the target class and $m$ the number of attributes. Hyperparameters $k$ and $l$ rounded to the nearest integer in the range $[1, n-1]$.}
\label{tab_optimal_parameter_values}
\begin{tabular}{lll}
\toprule
Data descriptor & Hyperparameter & Optimal default value \\
\midrule
NND  & $k$   & $1$\\
LNND & $k$   & $3.4\log n$\\
LOF  & $k$   & $2.5\log n$\\
SVM  & $\nu$ & $0.20$\\
     & $c$  & $0.25m$\\
ALP  & $k$   & $5.5\log n$\\
     & $l$   & $6.0\log n$\\
\bottomrule
\end{tabular}
\end{table}

Figure \ref{fig_auroc} displays the average AUROC of the data descriptors with hyperparameters. It is clear that NND, LNND and LOF produce better overall results with the Manhattan distance than with the Euclidean distance. After applying the running mean, the AUROC graphs approximate 1- or 2-dimensional unimodal curves with a prominent global maximum. For SVM and ALP, the mean weighted AUROC decreases quite slowly away from the global maximum, which means that these data descriptors are quite robust to small small changes in their hyperparameter values. On the basis of these graphs, we can recommend the default hyperparameter values listed in Table \ref{tab_optimal_parameter_values}. To avoid unwarranted precision, we have chosen to only determine these values up to multiples of $0.1$ (LNND and LOF), $0.05$ (SVM), and $0.5$ (ALP).

\begin{table}
\centering
\caption{Weighted mean rank, AUROC and standard deviation of AUROC across cross-validation folds (CVSD) of data descriptors.}
\label{tab_summary_stats}
\begin{tabular}{lrrrr}
\toprule
Data descriptor & Rank & AUROC &   CVSD \\
\midrule
            ALP & 3.51 & 0.817 & 0.0378 \\
            SVM & 3.90 & 0.814 & 0.0404 \\
            NND & 4.33 & 0.805 & 0.0414 \\
            LOF & 4.67 & 0.803 & 0.0417 \\
             MD & 4.68 & 0.806 & 0.0370 \\
             IF & 5.15 & 0.800 & 0.0479 \\
            EIF & 5.59 & 0.790 & 0.0504 \\
           LNND & 6.55 & 0.781 & 0.0438 \\
            SAE & 6.63 & 0.757 & 0.0539 \\
\bottomrule
\end{tabular}
\end{table}

The full leave-one-dataset-out AUROC values of the data descriptors are listed in Table \ref{tab_default_auroc}. The weighted mean rank and AUROC\footnote{Note that for ALP, SVM, NND, LOF and LNND, the weighted mean AUROC is lower than the highest scores in Figure \ref{fig_auroc}, since, in general, the leave-one-dataset-out hyperparameter values are not equal to the optimal values calculated on the basis of all datasets.} of the data descriptors are summarised in Table \ref{tab_summary_stats}. Because the AUROC for each dataset and target class is calculated through cross-validation, we can also measure the standard deviation across folds to gain an idea to what extend the performance is influenced by random factors. Generally speaking, we find that the lower-ranked data descriptors are more susceptible to statistical variation. We might reasonably expect to be able to improve their performance by making them more robust to statistical variation. The principal exceptions to this trend are MD and LNND, which show relatively low statistical variance, and this suggests a more limited potential for improvement.

\begin{table}
\centering
\caption{One-sided $p$-values of clustered Wilcoxon signed-rank tests of the hypotheses that the data descriptor in the row is ranked higher than the data descriptor in the column, with Holm-Bonferroni family-wise error correction applied to each row.}
\label{tab_p_values}
\begin{tabular}{llllllllll}
\toprule
      &    SVM &     NND &        LOF &       MD &      IF &        EIF &       LNND &        SAE \\
\midrule
  ALP & $0.14$ & $0.018$ & $< 0.0001$ &  $0.035$ & $0.035$ &  $0.00076$ & $< 0.0001$ & $< 0.0001$ \\
  SVM &        &  $0.51$ &    $0.029$ &  $0.050$ & $0.042$ & $< 0.0001$ & $< 0.0001$ & $< 0.0001$ \\
  NND &        &         &   $\geq 1$ &   $0.47$ &  $0.38$ &    $0.017$ & $< 0.0001$ & $< 0.0001$ \\
  LOF &        &         &            & $\geq 1$ &  $0.83$ &    $0.036$ & $< 0.0001$ &  $0.00028$ \\
   MD &        &         &            &          &  $0.35$ &     $0.11$ &   $0.0066$ & $< 0.0001$ \\
   IF &        &         &            &          &         &   $\geq 1$ &     $0.20$ &   $0.0068$ \\
  EIF &        &         &            &          &         &            &     $0.78$ &    $0.025$ \\
 LNND &        &         &            &          &         &            &            &     $0.27$ \\
\bottomrule
\end{tabular}
\end{table}

Next, we test which data descriptors can be said with certainty to perform better than others on the type of one-class classification problems represented by our sample. Table \ref{tab_p_values} contains the p-values of the one-sided clustered Wilcoxon signed-rank tests, corrected for family-wise error with the Holm-Bonferroni method.

We find that there is strong evidence that ALP outperforms the other data descriptors, except SVM, for which there is only weak evidence ($p = 0.14$). In particular, these results confirm ($p < 0.0001$) our hypothesis that ALP improves upon LOF and LNND. Among the other data descriptors, there is good evidence that SVM performs better than the others, except NND and MD (for which the evidence is weaker: $p = 0.050$), as well as strong evidence that LOF and NND outperform EIF, LNND and SAE, that MD outperforms LNND and SAE, and that IF and EIF outperform SAE, and weak evidence that MD performs better than EIF ($p = 0.11$).

We note that the performance of IF is the hardest to pin down: in fact we can only say with limited certainty that ALP ($p = 0.035$) and SVM ($p = 0.042$) perform better. In particular, we find no evidence that EIF performs better than IF. This may be because the type of dataset that motivated the formulation of EIF is not represented in our sample.

\begin{figure}
\centering
\includegraphics[width=\linewidth]{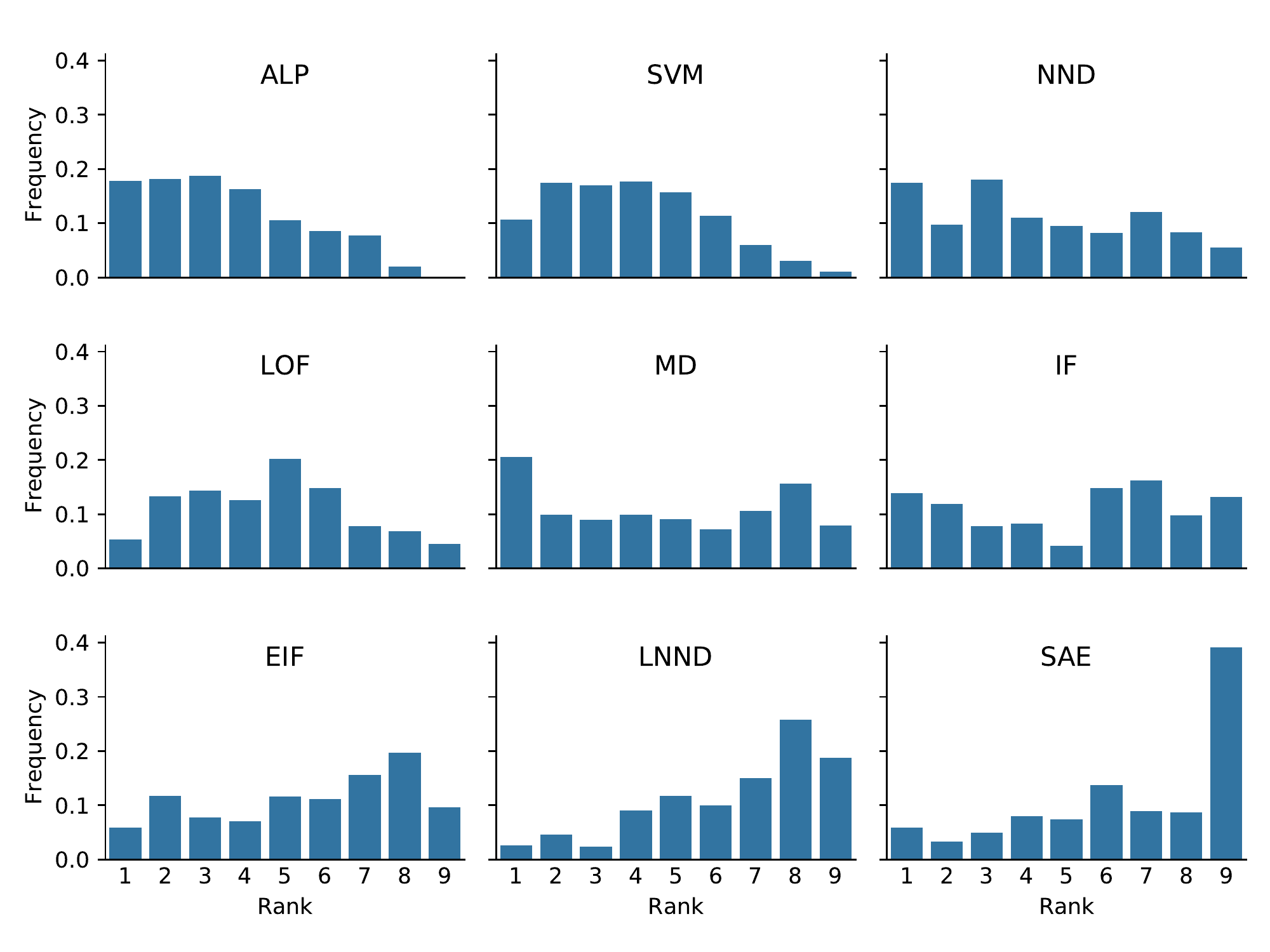}%
\caption{Average frequency ($y$-axis) of ranks ($x$-axis) of data descriptors, weighted by dataset, with ties distributed evenly among the respective ranks.}
\label{fig_ranks}
\end{figure}

To gain more insight into the typical performance of the data descriptors, we have summarised the frequency with which a data descriptor achieves a certain rank in Figure \ref{fig_ranks}. For ALP, SVM, LOF and LNND, the distributions of these frequencies are close to unimodal. In contrast, NND, MD, IF and EIF are less consistent --- their performance is good or bad as or more often than it is mediocre. The poor overall performance of SAE is reflected in the fact that it is the worst-ranked data descriptor for nearly 40\% of one-class classification problems. Yet it is also the best-ranked data descriptor in around 5\%.

\begin{figure}
\centering
\includegraphics[width=.9\linewidth]{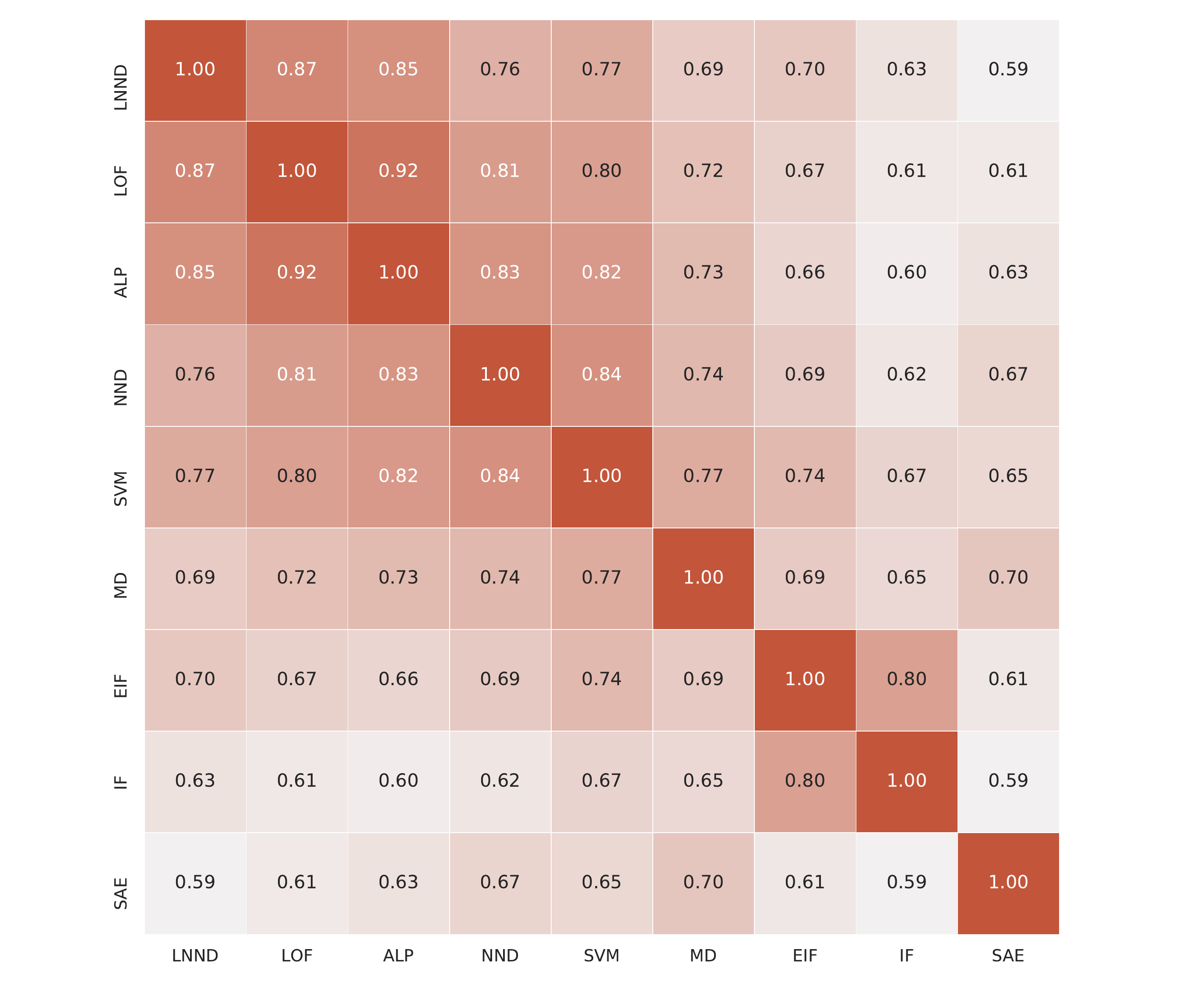}%
\caption{Weighted Kendall's $\tau$ between data descriptors.}
\label{fig_kendall}
\end{figure}

Figure \ref{fig_kendall} shows the extent to which the performances of the data descriptors correlate, as reflected by the weighted Kendall's $\tau$ \cite{vigna15weighted}. The strongest correlation can be found between the nearest neighbour based data descriptors and SVM, and between IF and EIF. This information is potentially useful when constructing ensembles of data descriptors. In addition to selecting data descriptors with good individual performance, it may be desirable to select pairs of data descriptors with low correlation that may be expected to complement (e.g. ALP and IF) rather than reinforce (e.g. ALP and LOF) each other. Indeed, when comparing all pairs of data descriptors on the basis of the maximum of their respective AUROC scores for each one-class classification problem, we find that the five best performing pairs are (ALP, MD), (NND, MD), (ALP, NND), (ALP, IF) and (NND, IF).

\begin{figure}
\centering
\includegraphics[width=\linewidth]{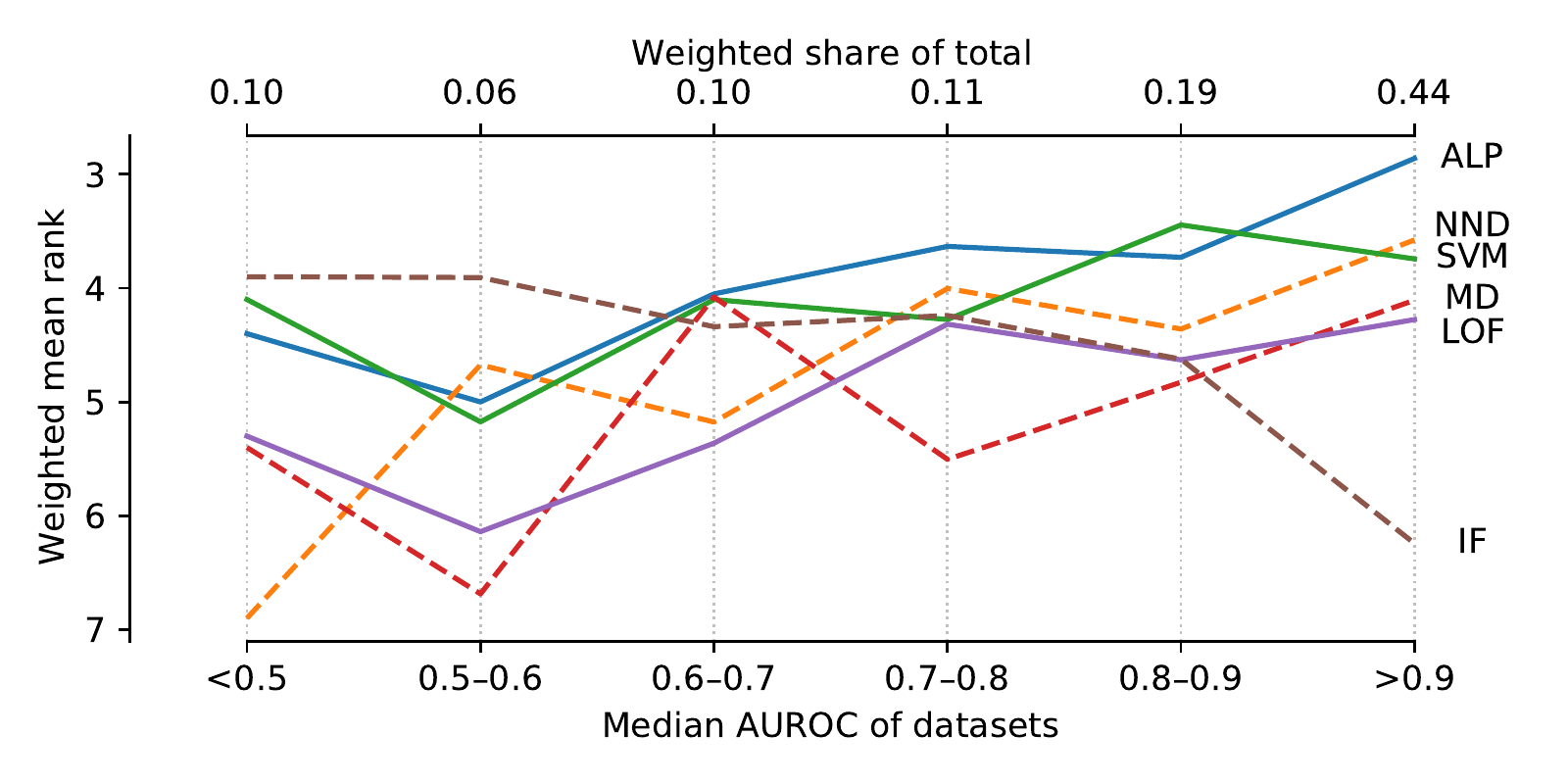}%
\caption{Weighted mean rank ($y$-axis) of selected data descriptors, stratified by the median AUROC of datasets across all data descriptors ($x$-axis).}
\label{fig_ranks_by_difficulty}
\end{figure}

There are a number of factors that potentially explain the performance of the data descriptors: the number of instances $n$, the dimensionality $m$ and the sparsity of the target data. In addition, we can use the median AUROC obtained by the data descriptors in this paper as an indication of the `simplicity' or `separability' of the evaluation data.

When we fit weighted linear regression models on the ranks of the data descriptors in terms of $\log n$, $\log m$, sparsity and median AUROC, we find that median AUROC is a positive significant factor for NND, LOF, ALP and MD, and a negative significant factor for LNND, IF and SAE. Surprisingly, there is no strong evidence that sparsity ($p = 0.400$) or dimensionality ($p = 0.802$) are positive factors for SAE. Unsurprisingly, given that SAE is a neural network, size \e{is} a significant positive factor ($p = 0.002$), with each order of magnitude (base $e$) increasing the rank of SAE by 0.31 places ($\pm 0.19$). Among the other data descriptors, sparsity may be a negative factor for LOF ($p = 0.058$).

The effect of median AUROC is illustrated in Figure \ref{fig_ranks_by_difficulty}. We see that the performance of ALP and SVM is broadly similar, except for the evaluation data that is the easiest to separate. This is the largest stratum in our sample, and it is the stratum for which ALP has the biggest relative advantage. ALP also has a fairly constant advantage over LOF across all strata. Interestingly, while IF performs suboptimally on evaluation data that is easily separable, it is competitive on harder problems.

\begin{figure}
\centering\subfloat[Construction times]{\includegraphics[width=.48\linewidth]{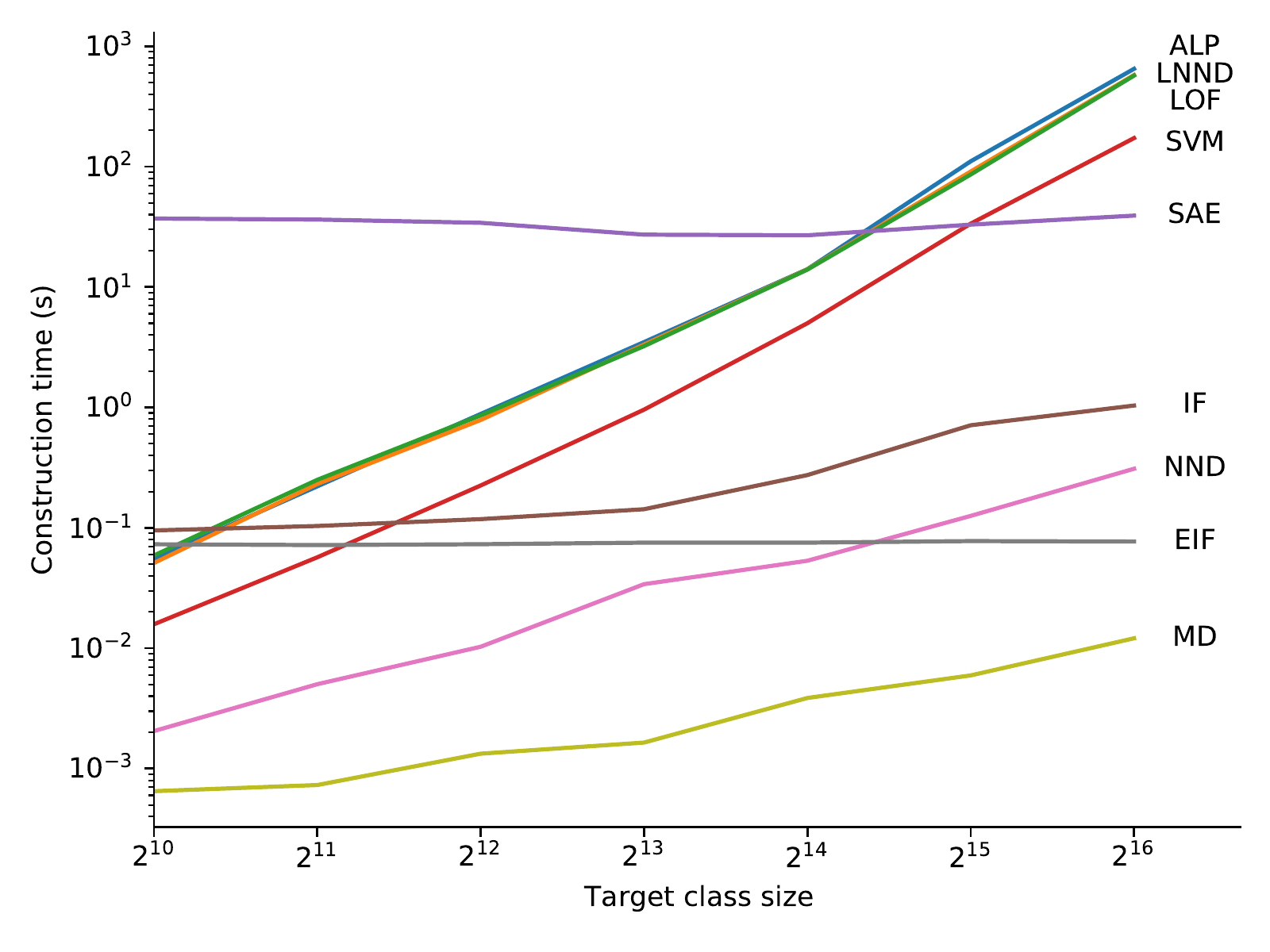}%
\label{fig_construction_times}}
\hfil
\subfloat[Query times]{\includegraphics[width=.48\linewidth]{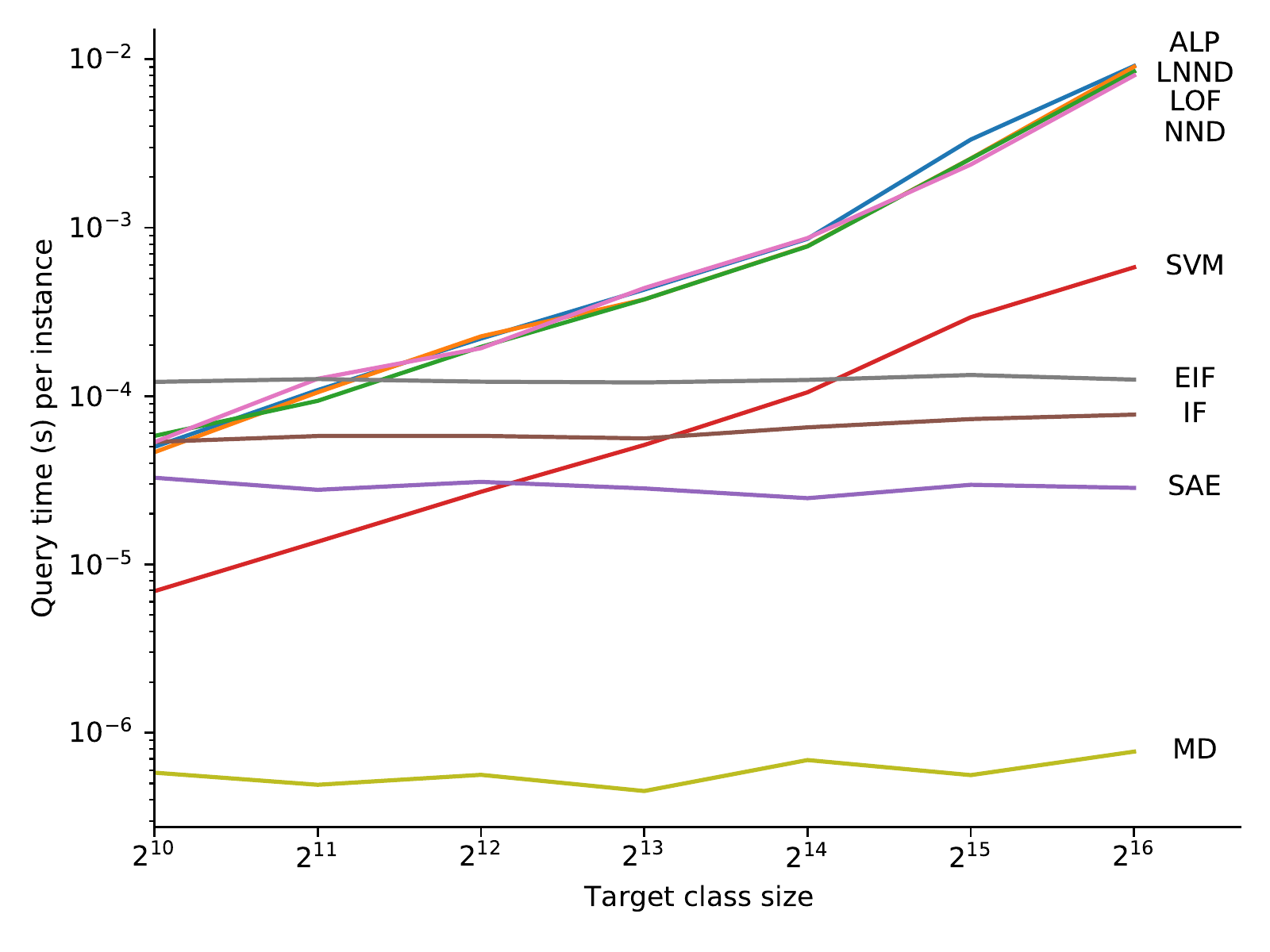}%
\label{fig_query_times}}\\
\caption{Mean construction and query times of data descriptors across five runs. Query times calculated on the basis of $2^{10}$ instances.}
\label{fig_run_times}
\end{figure}

Figure \ref{fig_run_times} contains the construction and query times of the data descriptors. The absolute times are implementation-dependent, but the way these times scale is nevertheless informative. There is a clear division between the nearest neighbour based data descriptors and SVM on the one hand, for which query times grow polynomially with training set size, and the other data descriptors, which keep query times near-constant. The query times of NND, LNND, LOF and ALP are near-identical, since they all revolve around a nearest neighbour query. The faster construction times of NND reflect the fact we only pre-calculate local distances for LNND, LOF and ALP. The construction times for SAE start high, but remain effectively constant, although this is dependent on the precise choice of early-stopping criteria. Similarly, the construction times of IF and EIF are constrained by the constant number of trees and number of samples per tree.

When comparing classification accuracy and computational performance, we see that the data descriptors present different possible compromises. ALP and SVM offer the highest accuracy, but slow run times with large datasets, whereas IF, MD and SAE offer much better computational performance with large datasets but lower accuracy. We note that in the case of ALP and SVM, this trade-off can be further tweaked by incorporating approximative nearest neighbour search and support vector machine algorithms.

\section{Conclusion}
\label{sec_conclusion}
This article has produced three main results. First, motivated by a number of perceived shortcomings in the existing data descriptors LNND and LOF, we have proposed Average Localised Proximity (ALP). Second, on the basis of a large collection of one-class classification problems derived from real-world datasets, we have determined optimal default hyperparameter values for ALP, as well as the existing data descriptors NND, LNND, LOF and SVM. And third, we have evaluated the performance of these data descriptors through a leave-one-dataset-out procedure, and found weak evidence that ALP outperforms SVM, and strong evidence that it outperforms all other data descriptors, in particular LNND and LOF.

In the Introduction, we pointed out that the data descriptors MD, IF, EIF and SAE have no user-tunable hyperparameters and may therefore appear simpler to apply, and even preferable when no data is available for tuning. The optimal default hyperparameter values identified in this article allow all of the data descriptors discussed here to also be used as ``easy-to-use black-box method for practitioners'', as called for by Schölkopf et al.~\cite{scholkopf99estimating}. Moreover, we have shown that with their default hyperparameter values, ALP and SVM generally perform better than MD, IF, EIF and SAE. The ability to further optimise these hyperparameters for specific datasets should therefore be seen as an additional advantage, not a necessary requirement.

We see a number of avenues for further research. Firstly, it may be possible to further reparametrise some hyperparameters in a way that would allow the identification of even better performing default values. Secondly, we set out to evaluate general data descriptor performance using a variety of different one-class classification problems, but it may also be worthwhile to investigate whether certain data descriptors are better suited to specific types of problems. Thirdly, we have evaluated novelty detection, and it remains to be seen whether the results in this article can be reproduced for unsupervised outlier detection. Fourthly, given the good default performance of ALP, its performance should also be compared to SVM in a setting with per-dataset hyperparameter tuning. And lastly, given the better computational performance of MD, IF, EIF and SAE, we are interested in exploring the trade-off between accuracy and run time --- on the one hand, ALP and SVM can be sped up by incorporating approximative algorithms, while on the other hand, it may be possible to modify the IF, EIF and SAE algorithms to improve their accuracy, at some computational cost.

\section*{Acknowledgements}
\label{sec_acknowledgement}
The authors would like to thank James McDermott for a number of useful comments and suggestions.

The research reported in this paper was conducted with the financial support of the Odysseus programme of the Research Foundation -- Flanders (FWO). D. Peralta is a Postdoctoral Fellow of the Research Foundation -- Flanders (FWO, 170303/12X1619N).

\appendix

\section{Full results}
\setlength{\LTleft}{-20cm plus -1fill}
\setlength{\LTright}{\LTleft}
{\scriptsize
\begin{longtable}{lrlrrrrrrrrrrr}
\caption{AUROC of one-class classifiers with default parameter values.}\label{tab_default_auroc}\\
\toprule
Dataset & $m$ & Class & $n$ & $s$ & NND & LNND & LOF & MD & SVM & IF & EIF & SAE & ALP\\
\midrule
\endhead
\midrule
\multicolumn{14}{r}{{Continued on next page}} \\
\midrule
\endfoot

\bottomrule
\endlastfoot
appendicitis & 7   & 0 & 85     & 0.05 &         0.741 &         0.644 &         0.744 &         0.691 &         0.665 &  \bftab 0.765 &         0.734 &         0.518 &         0.721 \\
      &     & 1 & 21     & 0.12 &         0.688 &         0.741 &         0.787 &         0.683 &         0.686 &  \bftab 0.799 &         0.787 &         0.603 &         0.780 \\
\midrule
avila & 10  & A & 8572   & 0.08 &  \bftab 0.914 &         0.759 &         0.844 &         0.622 &         0.743 &         0.599 &         0.596 &         0.582 &         0.891 \\
      &     & B & 10     & 0.57 &  \bftab 1.000 &  \bftab 1.000 &  \bftab 1.000 &         0.844 &         1.000 &         0.910 &  \bftab 1.000 &         0.894 &  \bftab 1.000 \\
      &     & C & 206    & 0.14 &  \bftab 0.964 &         0.825 &         0.944 &         0.591 &         0.923 &         0.631 &         0.706 &         0.566 &         0.959 \\
      &     & D & 705    & 0.11 &  \bftab 0.974 &         0.887 &         0.959 &         0.692 &         0.952 &         0.665 &         0.688 &         0.685 &         0.966 \\
      &     & E & 2190   & 0.09 &  \bftab 0.967 &         0.828 &         0.934 &         0.567 &         0.857 &         0.649 &         0.672 &         0.569 &         0.953 \\
      &     & F & 3923   & 0.09 &  \bftab 0.936 &         0.785 &         0.878 &         0.698 &         0.804 &         0.698 &         0.697 &         0.634 &         0.912 \\
      &     & G & 893    & 0.13 &  \bftab 0.987 &         0.889 &         0.962 &         0.789 &         0.959 &         0.762 &         0.844 &         0.803 &         0.981 \\
      &     & H & 1039   & 0.12 &         0.973 &         0.884 &         0.951 &         0.871 &         0.929 &         0.840 &         0.847 &         0.739 &  \bftab 0.974 \\
      &     & I & 1663   & 0.10 &  \bftab 0.996 &         0.915 &         0.980 &         0.799 &         0.994 &         0.837 &         0.852 &         0.900 &         0.982 \\
      &     & W & 89     & 0.19 &  \bftab 0.969 &         0.788 &         0.954 &         0.958 &         0.896 &         0.627 &         0.801 &         0.920 &         0.950 \\
      &     & X & 1044   & 0.09 &  \bftab 0.983 &         0.876 &         0.972 &         0.732 &         0.960 &         0.699 &         0.678 &         0.751 &         0.981 \\
      &     & Y & 533    & 0.08 &  \bftab 0.998 &         0.948 &         0.990 &         0.885 &         0.994 &         0.840 &         0.861 &         0.808 &         0.995 \\
\midrule
banknote & 4   & 0 & 762    & 0.01 &         0.998 &         0.946 &         0.993 &  \bftab 0.999 &         0.997 &         0.922 &         0.950 &         0.985 &         0.996 \\
      &     & 1 & 610    & 0.00 &  \bftab 1.000 &         0.993 &         0.999 &         0.994 &         0.997 &         0.884 &         0.941 &         0.999 &         1.000 \\
\midrule
biodeg & 41  & NRB & 699    & 0.46 &  \bftab 0.597 &         0.557 &         0.502 &         0.496 &         0.558 &         0.443 &         0.587 &         0.516 &         0.538 \\
      &     & RB & 356    & 0.51 &         0.852 &         0.784 &         0.850 &  \bftab 0.870 &         0.835 &         0.866 &         0.758 &         0.847 &         0.866 \\
\midrule
breasttissue & 9   & adi & 22     & 0.06 &         0.925 &         0.904 &         0.920 &         0.893 &         0.926 &         0.816 &         0.881 &  \bftab 0.947 &         0.928 \\
      &     & car & 21     & 0.05 &         0.926 &         0.939 &  \bftab 0.972 &         0.924 &         0.945 &         0.886 &         0.910 &         0.830 &         0.962 \\
      &     & con & 14     & 0.08 &         0.888 &         0.858 &         0.933 &         0.836 &         0.874 &         0.778 &         0.817 &         0.841 &  \bftab 0.936 \\
      &     & fad & 15     & 0.07 &  \bftab 0.896 &         0.847 &         0.851 &         0.862 &         0.888 &         0.814 &         0.858 &         0.822 &         0.854 \\
      &     & gla & 16     & 0.12 &         0.922 &         0.866 &         0.918 &  \bftab 0.939 &         0.904 &         0.910 &         0.912 &         0.831 &         0.919 \\
      &     & mas & 18     & 0.06 &         0.737 &         0.739 &         0.696 &  \bftab 0.741 &         0.715 &         0.728 &         0.661 &         0.672 &         0.707 \\
\midrule
coimbra & 9   & 1 & 52     & 0.04 &  \bftab 0.743 &         0.691 &         0.714 &         0.665 &         0.732 &         0.639 &         0.706 &         0.632 &         0.714 \\
      &     & 2 & 64     & 0.04 &         0.424 &         0.417 &         0.479 &  \bftab 0.587 &         0.520 &         0.548 &         0.441 &         0.561 &         0.491 \\
\midrule
column & 6   & DH & 60     & 0.03 &         0.875 &         0.837 &         0.856 &         0.874 &  \bftab 0.888 &         0.846 &         0.857 &         0.801 &         0.876 \\
      &     & NO & 100    & 0.02 &         0.875 &         0.875 &  \bftab 0.881 &         0.869 &         0.870 &         0.871 &         0.880 &         0.866 &         0.880 \\
      &     & SL & 150    & 0.01 &         0.742 &         0.749 &         0.813 &         0.712 &  \bftab 0.847 &         0.817 &         0.772 &         0.523 &         0.846 \\
\midrule
debrecen & 19  & 0 & 540    & 0.28 &         0.619 &         0.642 &         0.636 &  \bftab 0.768 &         0.631 &         0.678 &         0.622 &         0.660 &         0.643 \\
      &     & 1 & 611    & 0.25 &         0.404 &         0.448 &         0.432 &         0.370 &         0.386 &         0.400 &         0.388 &         0.415 &  \bftab 0.450 \\
\midrule
dermatology & 34  & 1 & 111    & 0.70 &         0.995 &         0.953 &         0.985 &         0.955 &  \bftab 1.000 &         0.985 &         0.992 &         0.981 &         0.996 \\
      &     & 2 & 60     & 0.79 &         0.952 &         0.938 &         0.953 &         0.937 &         0.959 &         0.878 &         0.952 &         0.930 &  \bftab 0.961 \\
      &     & 3 & 71     & 0.74 &         0.998 &         0.990 &         0.997 &  \bftab 0.999 &         0.999 &         0.996 &         0.997 &         0.958 &         0.997 \\
      &     & 4 & 48     & 0.83 &         0.985 &         0.974 &  \bftab 0.986 &         0.961 &         0.976 &         0.883 &         0.964 &         0.972 &         0.986 \\
      &     & 5 & 48     & 0.79 &         0.997 &         0.982 &  \bftab 0.999 &         0.975 &         0.997 &         0.955 &         0.996 &         0.948 &  \bftab 0.999 \\
      &     & 6 & 20     & 0.75 &         0.993 &         0.981 &  \bftab 0.998 &         0.997 &         0.995 &         0.962 &         0.932 &         0.734 &  \bftab 0.998 \\
\midrule
divorce & 54  & 0 & 86     & 0.69 &  \bftab 1.000 &         0.995 &  \bftab 1.000 &         0.992 &         0.998 &  \bftab 1.000 &         0.997 &         0.936 &  \bftab 1.000 \\
      &     & 1 & 84     & 0.55 &         0.888 &         0.873 &         0.852 &         0.889 &         0.900 &  \bftab 0.999 &         0.994 &         0.819 &         0.878 \\
\midrule
ecoli & 7   & cp & 143    & 0.33 &         0.968 &         0.966 &  \bftab 0.979 &         0.974 &         0.966 &         0.972 &         0.969 &         0.951 &         0.978 \\
      &     & im & 77     & 0.34 &         0.869 &         0.831 &         0.886 &         0.894 &         0.885 &  \bftab 0.906 &         0.890 &         0.730 &         0.891 \\
      &     & imU & 35     & 0.36 &         0.908 &         0.892 &         0.910 &         0.910 &         0.913 &  \bftab 0.935 &         0.933 &         0.758 &         0.916 \\
      &     & om & 20     & 0.38 &         0.949 &         0.963 &         0.972 &         0.967 &         0.962 &         0.972 &         0.960 &         0.562 &  \bftab 0.979 \\
      &     & omL & 5      & 0.46 &         0.985 &  \bftab 0.991 &         0.983 &         0.985 &         0.985 &         0.632 &         0.655 &         0.852 &         0.988 \\
      &     & pp & 52     & 0.36 &         0.925 &         0.931 &         0.951 &         0.928 &         0.938 &         0.938 &         0.936 &         0.813 &  \bftab 0.952 \\
\midrule
faults & 27  & Bumps & 402    & 0.14 &  \bftab 0.841 &         0.777 &         0.823 &         0.826 &         0.834 &         0.821 &         0.792 &         0.777 &         0.833 \\
      &     & Dirtiness & 55     & 0.21 &  \bftab 0.947 &         0.824 &         0.888 &         0.931 &         0.920 &         0.886 &         0.876 &         0.930 &         0.944 \\
      &     & K\_Scatch & 391    & 0.22 &  \bftab 0.986 &         0.804 &         0.892 &         0.983 &         0.983 &         0.972 &         0.972 &         0.967 &         0.904 \\
      &     & Other\_Faults & 673    & 0.13 &  \bftab 0.662 &         0.585 &         0.632 &         0.600 &         0.608 &         0.603 &         0.597 &         0.620 &         0.631 \\
      &     & Pastry & 158    & 0.17 &         0.854 &         0.779 &         0.815 &  \bftab 0.870 &         0.837 &         0.852 &         0.786 &         0.840 &         0.840 \\
      &     & Stains & 72     & 0.31 &         0.997 &         0.995 &  \bftab 0.997 &         0.994 &         0.986 &         0.994 &         0.994 &         0.981 &         0.997 \\
      &     & Z\_Scratch & 190    & 0.19 &         0.972 &         0.932 &         0.936 &         0.958 &  \bftab 0.975 &         0.915 &         0.946 &         0.922 &         0.971 \\
\midrule
foresttypes & 27  & d & 159    & 0.04 &         0.838 &         0.836 &         0.870 &         0.844 &         0.868 &         0.880 &         0.818 &         0.776 &  \bftab 0.880 \\
      &     & h & 86     & 0.09 &         0.936 &         0.919 &         0.938 &         0.898 &         0.941 &         0.937 &  \bftab 0.950 &         0.885 &         0.937 \\
      &     & o & 83     & 0.04 &         0.690 &         0.736 &         0.732 &         0.698 &         0.739 &  \bftab 0.879 &         0.770 &         0.780 &         0.713 \\
      &     & s & 195    & 0.06 &         0.931 &         0.907 &         0.941 &         0.910 &         0.936 &         0.940 &         0.928 &         0.825 &  \bftab 0.944 \\
\midrule
glass & 9   & 1 & 70     & 0.23 &         0.803 &         0.820 &         0.874 &         0.797 &         0.840 &         0.800 &         0.795 &         0.813 &  \bftab 0.893 \\
      &     & 2 & 76     & 0.21 &         0.689 &         0.641 &         0.708 &         0.659 &         0.674 &         0.637 &         0.623 &         0.568 &  \bftab 0.718 \\
      &     & 3 & 17     & 0.27 &         0.719 &         0.715 &         0.689 &  \bftab 0.845 &         0.806 &         0.661 &         0.734 &         0.753 &         0.776 \\
      &     & 5 & 13     & 0.32 &         0.641 &         0.571 &         0.611 &         0.702 &         0.762 &         0.704 &         0.610 &         0.385 &  \bftab 0.763 \\
      &     & 6 & 9      & 0.43 &         0.668 &         0.563 &         0.571 &  \bftab 0.971 &         0.599 &         0.456 &         0.524 &         0.654 &         0.829 \\
      &     & 7 & 29     & 0.29 &         0.801 &         0.814 &         0.800 &         0.738 &  \bftab 0.865 &         0.827 &         0.794 &         0.750 &         0.839 \\
\midrule
haberman & 3   & 1 & 225    & 0.22 &         0.679 &         0.642 &         0.625 &         0.616 &         0.652 &         0.670 &  \bftab 0.695 &         0.561 &         0.608 \\
      &     & 2 & 81     & 0.16 &         0.404 &         0.458 &         0.464 &         0.515 &  \bftab 0.526 &         0.456 &         0.438 &         0.507 &         0.469 \\
\midrule
htru2 & 8   & 0 & 16259  & 0.00 &  \bftab 0.953 &         0.918 &         0.952 &         0.944 &         0.951 &         0.949 &         0.937 &         0.946 &         0.946 \\
      &     & 1 & 1639   & 0.00 &         0.837 &         0.667 &         0.684 &         0.870 &         0.913 &         0.900 &  \bftab 0.924 &         0.750 &         0.677 \\
\midrule
ilpd & 10  & 1 & 414    & 0.17 &         0.346 &         0.371 &         0.342 &  \bftab 0.453 &         0.365 &         0.400 &         0.345 &         0.428 &         0.389 \\
      &     & 2 & 165    & 0.19 &  \bftab 0.753 &         0.725 &         0.729 &         0.703 &         0.753 &         0.705 &         0.738 &         0.697 &         0.734 \\
\midrule
ionosphere & 34  & b & 126    & 0.32 &         0.363 &  \bftab 0.584 &         0.426 &         0.246 &         0.303 &         0.367 &         0.301 &         0.300 &         0.370 \\
      &     & g & 225    & 0.13 &         0.949 &         0.903 &         0.951 &         0.963 &  \bftab 0.971 &         0.908 &         0.956 &         0.947 &         0.956 \\
\midrule
iris & 4   & Iris-setosa & 50     & 0.29 &  \bftab 1.000 &  \bftab 1.000 &  \bftab 1.000 &  \bftab 1.000 &  \bftab 1.000 &  \bftab 1.000 &  \bftab 1.000 &  \bftab 1.000 &  \bftab 1.000 \\
      &     & Iris-versicolor & 50     & 0.17 &         0.974 &         0.962 &  \bftab 0.990 &  \bftab 0.990 &         0.975 &         0.979 &  \bftab 0.990 &         0.968 &         0.985 \\
      &     & Iris-virginica & 50     & 0.19 &         0.942 &         0.916 &         0.942 &  \bftab 0.962 &         0.955 &         0.941 &         0.924 &         0.925 &         0.956 \\
\midrule
landsat & 36  & 1 & 1533   & 0.10 &  \bftab 0.997 &         0.966 &         0.991 &         0.969 &         0.992 &         0.990 &         0.992 &         0.937 &         0.992 \\
      &     & 2 & 703    & 0.13 &  \bftab 0.988 &         0.747 &         0.860 &         0.859 &         0.986 &         0.986 &         0.975 &         0.916 &         0.881 \\
      &     & 3 & 1358   & 0.14 &         0.975 &         0.939 &         0.971 &         0.908 &  \bftab 0.976 &         0.969 &         0.970 &         0.928 &         0.974 \\
      &     & 4 & 626    & 0.12 &         0.896 &         0.812 &         0.843 &         0.799 &         0.888 &  \bftab 0.925 &         0.902 &         0.792 &         0.853 \\
      &     & 5 & 707    & 0.09 &  \bftab 0.941 &         0.779 &         0.858 &         0.760 &         0.924 &         0.893 &         0.898 &         0.792 &         0.894 \\
      &     & 7 & 1508   & 0.12 &         0.954 &         0.872 &         0.903 &         0.894 &         0.940 &         0.960 &  \bftab 0.960 &         0.910 &         0.903 \\
\midrule
leaf & 14  & 1 & 12     & 0.12 &         0.980 &         0.978 &  \bftab 0.981 &         0.978 &         0.978 &         0.946 &         0.915 &         0.888 &         0.978 \\
      &     & 2 & 10     & 0.16 &         0.959 &         0.953 &         0.964 &         0.932 &         0.923 &  \bftab 0.980 &         0.977 &         0.871 &         0.958 \\
      &     & 3 & 10     & 0.13 &         0.968 &         0.964 &         0.986 &         0.977 &         0.974 &         0.889 &         0.885 &         0.780 &  \bftab 0.991 \\
      &     & 4 & 8      & 0.14 &         0.866 &         0.826 &         0.834 &         0.887 &         0.891 &         0.703 &         0.837 &         0.775 &  \bftab 0.907 \\
      &     & 5 & 12     & 0.08 &         0.998 &         0.993 &  \bftab 0.999 &         0.955 &         0.995 &         0.956 &         0.994 &         0.795 &         0.997 \\
      &     & 6 & 8      & 0.13 &  \bftab 0.997 &         0.977 &         0.989 &         0.967 &  \bftab 0.997 &         0.921 &         0.931 &         0.721 &         0.991 \\
      &     & 7 & 10     & 0.11 &         0.961 &         0.964 &         0.962 &  \bftab 0.985 &         0.971 &         0.867 &         0.958 &         0.742 &         0.974 \\
      &     & 8 & 11     & 0.11 &  \bftab 1.000 &  \bftab 1.000 &  \bftab 1.000 &         0.997 &         0.966 &         0.921 &         0.997 &         0.784 &  \bftab 1.000 \\
      &     & 9 & 14     & 0.08 &         0.957 &         0.921 &         0.949 &  \bftab 0.963 &         0.958 &         0.895 &         0.928 &         0.906 &         0.959 \\
      &     & 10 & 13     & 0.09 &         0.995 &         0.993 &  \bftab 0.996 &         0.946 &         0.993 &         0.971 &         0.952 &         0.741 &         0.995 \\
      &     & 11 & 16     & 0.07 &         0.999 &         0.998 &  \bftab 1.000 &         0.996 &  \bftab 1.000 &         0.960 &         0.998 &         0.890 &         0.999 \\
      &     & 12 & 12     & 0.09 &         0.970 &         0.953 &         0.966 &         0.898 &  \bftab 0.974 &         0.944 &         0.939 &         0.721 &         0.964 \\
      &     & 13 & 13     & 0.13 &         0.977 &         0.981 &  \bftab 0.986 &         0.963 &         0.981 &         0.978 &         0.973 &         0.870 &         0.983 \\
      &     & 14 & 12     & 0.10 &  \bftab 0.948 &         0.948 &         0.938 &         0.876 &         0.914 &         0.904 &         0.935 &         0.877 &         0.937 \\
      &     & 15 & 10     & 0.11 &  \bftab 1.000 &         0.998 &  \bftab 1.000 &         0.988 &  \bftab 1.000 &         0.977 &         0.985 &         0.850 &  \bftab 1.000 \\
      &     & 22 & 12     & 0.09 &         0.793 &         0.811 &         0.804 &  \bftab 0.880 &         0.807 &         0.754 &         0.769 &         0.754 &         0.845 \\
      &     & 23 & 11     & 0.09 &         0.996 &         0.997 &         0.995 &         0.983 &  \bftab 0.997 &         0.807 &         0.972 &         0.845 &  \bftab 0.997 \\
      &     & 24 & 13     & 0.10 &         0.933 &         0.904 &         0.922 &  \bftab 0.952 &         0.920 &         0.738 &         0.877 &         0.740 &         0.939 \\
      &     & 25 & 9      & 0.13 &         0.998 &         0.998 &  \bftab 1.000 &         0.977 &         0.998 &         0.932 &         0.954 &         0.905 &         0.998 \\
      &     & 26 & 12     & 0.12 &         0.895 &         0.845 &         0.857 &  \bftab 0.930 &         0.894 &         0.839 &         0.819 &         0.873 &         0.881 \\
      &     & 27 & 11     & 0.15 &         0.968 &         0.883 &         0.918 &  \bftab 0.982 &         0.927 &         0.847 &         0.867 &         0.885 &         0.948 \\
      &     & 28 & 12     & 0.14 &         0.962 &         0.895 &         0.939 &  \bftab 0.974 &         0.967 &         0.918 &         0.913 &         0.850 &         0.954 \\
      &     & 29 & 12     & 0.11 &  \bftab 1.000 &         0.994 &  \bftab 1.000 &         0.995 &  \bftab 1.000 &         0.998 &         0.983 &         0.787 &  \bftab 1.000 \\
      &     & 30 & 12     & 0.10 &         0.992 &         0.989 &         0.995 &         0.990 &         0.995 &         0.913 &         0.974 &         0.863 &  \bftab 0.998 \\
      &     & 31 & 11     & 0.10 &  \bftab 0.999 &         0.996 &         0.997 &         0.997 &         0.998 &         0.922 &         0.937 &         0.911 &         0.997 \\
      &     & 32 & 11     & 0.12 &         0.962 &         0.959 &         0.966 &         0.956 &         0.965 &         0.852 &         0.956 &         0.814 &  \bftab 0.973 \\
      &     & 33 & 11     & 0.13 &         0.922 &         0.920 &         0.947 &         0.933 &         0.922 &  \bftab 0.958 &         0.927 &         0.792 &         0.950 \\
      &     & 34 & 11     & 0.10 &  \bftab 0.999 &         0.990 &         0.996 &         0.997 &         0.947 &         0.805 &         0.796 &         0.956 &         0.997 \\
      &     & 35 & 11     & 0.13 &         0.943 &  \bftab 0.945 &         0.938 &         0.903 &         0.882 &         0.918 &         0.876 &         0.797 &         0.943 \\
      &     & 36 & 10     & 0.10 &         0.994 &  \bftab 1.000 &  \bftab 1.000 &         0.994 &         0.992 &         0.945 &         0.947 &         0.879 &  \bftab 1.000 \\
\midrule
letter & 16  & A & 789    & 0.35 &         1.000 &         0.910 &         0.977 &         0.995 &  \bftab 1.000 &         0.941 &         0.948 &         0.965 &         0.980 \\
      &     & B & 766    & 0.35 &  \bftab 0.992 &         0.934 &         0.975 &         0.987 &         0.990 &         0.924 &         0.932 &         0.956 &         0.984 \\
      &     & C & 736    & 0.31 &  \bftab 0.996 &         0.916 &         0.972 &         0.981 &         0.993 &         0.935 &         0.942 &         0.945 &         0.978 \\
      &     & D & 805    & 0.35 &  \bftab 0.995 &         0.933 &         0.978 &         0.983 &         0.992 &         0.907 &         0.924 &         0.916 &         0.986 \\
      &     & E & 768    & 0.33 &  \bftab 0.988 &         0.927 &         0.981 &         0.980 &         0.973 &         0.876 &         0.885 &         0.939 &         0.986 \\
      &     & F & 775    & 0.27 &  \bftab 0.995 &         0.917 &         0.975 &         0.979 &         0.990 &         0.944 &         0.920 &         0.928 &         0.980 \\
      &     & G & 773    & 0.32 &  \bftab 0.991 &         0.910 &         0.967 &         0.974 &         0.977 &         0.890 &         0.900 &         0.927 &         0.978 \\
      &     & H & 734    & 0.39 &         0.954 &         0.856 &         0.951 &         0.936 &         0.928 &         0.844 &         0.823 &         0.907 &  \bftab 0.964 \\
      &     & I & 755    & 0.49 &  \bftab 0.996 &         0.829 &         0.899 &         0.978 &         0.991 &         0.942 &         0.941 &         0.970 &         0.946 \\
      &     & J & 747    & 0.28 &  \bftab 0.997 &         0.902 &         0.961 &         0.982 &         0.991 &         0.951 &         0.942 &         0.942 &         0.970 \\
      &     & K & 739    & 0.30 &  \bftab 0.984 &         0.924 &         0.974 &         0.969 &         0.962 &         0.870 &         0.869 &         0.949 &         0.983 \\
      &     & L & 761    & 0.28 &  \bftab 0.997 &         0.884 &         0.953 &         0.987 &         0.985 &         0.916 &         0.939 &         0.952 &         0.964 \\
      &     & M & 792    & 0.32 &  \bftab 0.995 &         0.906 &         0.967 &         0.972 &         0.993 &         0.908 &         0.886 &         0.947 &         0.985 \\
      &     & N & 783    & 0.32 &  \bftab 0.992 &         0.897 &         0.953 &         0.986 &         0.989 &         0.926 &         0.924 &         0.941 &         0.979 \\
      &     & O & 753    & 0.41 &  \bftab 0.990 &         0.933 &         0.986 &         0.986 &         0.990 &         0.922 &         0.934 &         0.956 &         0.990 \\
      &     & P & 803    & 0.30 &  \bftab 0.995 &         0.919 &         0.970 &         0.983 &         0.992 &         0.945 &         0.951 &         0.928 &         0.977 \\
      &     & Q & 783    & 0.30 &         0.985 &         0.941 &         0.988 &         0.976 &         0.975 &         0.877 &         0.884 &         0.938 &  \bftab 0.992 \\
      &     & R & 758    & 0.31 &  \bftab 0.993 &         0.952 &         0.985 &         0.983 &         0.986 &         0.930 &         0.934 &         0.939 &         0.989 \\
      &     & S & 748    & 0.32 &         0.980 &         0.902 &         0.978 &         0.966 &         0.973 &         0.835 &         0.828 &         0.932 &  \bftab 0.983 \\
      &     & T & 796    & 0.27 &  \bftab 0.996 &         0.924 &         0.973 &         0.987 &         0.985 &         0.944 &         0.959 &         0.947 &         0.983 \\
      &     & U & 813    & 0.34 &  \bftab 0.996 &         0.915 &         0.972 &         0.989 &         0.989 &         0.906 &         0.900 &         0.968 &         0.981 \\
      &     & V & 764    & 0.35 &  \bftab 0.997 &         0.937 &         0.976 &         0.993 &         0.995 &         0.949 &         0.958 &         0.978 &         0.985 \\
      &     & W & 752    & 0.32 &  \bftab 0.999 &         0.926 &         0.986 &         0.994 &         0.998 &         0.965 &         0.978 &         0.975 &         0.989 \\
      &     & X & 787    & 0.40 &         0.982 &         0.911 &         0.973 &         0.984 &         0.983 &         0.889 &         0.888 &         0.945 &  \bftab 0.984 \\
      &     & Y & 786    & 0.31 &  \bftab 0.994 &         0.886 &         0.970 &         0.977 &         0.983 &         0.922 &         0.936 &         0.962 &         0.976 \\
      &     & Z & 734    & 0.37 &  \bftab 0.991 &         0.895 &         0.975 &         0.977 &         0.988 &         0.900 &         0.899 &         0.954 &         0.988 \\
\midrule
magic & 10  & g & 12332  & 0.00 &         0.845 &         0.808 &         0.860 &         0.803 &         0.787 &         0.757 &         0.777 &         0.777 &  \bftab 0.873 \\
      &     & h & 6688   & 0.00 &         0.453 &         0.486 &         0.483 &         0.435 &  \bftab 0.491 &         0.442 &         0.406 &         0.455 &         0.484 \\
\midrule
mfeat & 649 & 0 & 200    & 0.30 &         0.998 &         0.983 &         0.997 &         0.996 &         0.996 &         0.988 &         0.983 &  \bftab 0.999 &         0.996 \\
      &     & 1 & 200    & 0.30 &         0.995 &         0.916 &         0.990 &         0.995 &         0.993 &         0.980 &         0.972 &         0.982 &  \bftab 0.997 \\
      &     & 2 & 200    & 0.29 &  \bftab 0.999 &         0.987 &         0.999 &         0.998 &         0.998 &         0.993 &         0.988 &         0.983 &         0.999 \\
      &     & 3 & 200    & 0.29 &  \bftab 0.985 &         0.933 &         0.976 &         0.983 &         0.983 &         0.973 &         0.970 &         0.967 &         0.979 \\
      &     & 4 & 200    & 0.29 &  \bftab 0.997 &         0.976 &         0.996 &         0.993 &         0.994 &         0.989 &         0.983 &         0.986 &         0.996 \\
      &     & 5 & 200    & 0.28 &  \bftab 0.989 &         0.930 &         0.985 &         0.984 &         0.981 &         0.967 &         0.960 &         0.936 &         0.983 \\
      &     & 6 & 200    & 0.29 &  \bftab 0.997 &         0.969 &         0.993 &         0.997 &         0.995 &         0.987 &         0.987 &         0.988 &         0.995 \\
      &     & 7 & 200    & 0.31 &  \bftab 0.999 &         0.982 &         0.998 &         0.997 &         0.998 &         0.990 &         0.987 &         0.995 &         0.998 \\
      &     & 8 & 200    & 0.27 &         0.987 &         0.953 &         0.984 &  \bftab 0.988 &         0.981 &         0.968 &         0.962 &         0.963 &         0.985 \\
      &     & 9 & 200    & 0.29 &         0.997 &         0.968 &         0.995 &  \bftab 0.997 &         0.996 &         0.986 &         0.983 &         0.982 &         0.995 \\
\midrule
miniboone & 50  & 0 & 93565  & 0.03 &         0.506 &         0.590 &         0.612 &         0.693 &         0.717 &         0.775 &         0.709 &  \bftab 0.829 &         0.600 \\
      &     & 1 & 36499  & 0.02 &         0.898 &         0.807 &  \bftab 0.902 &         0.865 &         0.842 &         0.799 &         0.784 &         0.700 &         0.901 \\
\midrule
new-thyroid & 5   & 1 & 150    & 0.08 &         0.972 &         0.957 &         0.991 &         0.987 &         0.988 &         0.992 &         0.987 &         0.965 &  \bftab 0.992 \\
      &     & 2 & 35     & 0.13 &         0.963 &         0.937 &         0.956 &         0.965 &         0.966 &         0.879 &         0.944 &         0.960 &  \bftab 0.971 \\
      &     & 3 & 30     & 0.11 &         0.808 &         0.779 &         0.863 &         0.938 &         0.851 &         0.918 &         0.785 &  \bftab 0.978 &         0.878 \\
\midrule
page-blocks & 10  & 1 & 4913   & 0.04 &         0.908 &         0.899 &         0.934 &         0.957 &  \bftab 0.964 &         0.924 &         0.930 &         0.925 &         0.940 \\
      &     & 2 & 329    & 0.19 &         0.916 &         0.869 &         0.916 &         0.898 &         0.924 &         0.875 &         0.916 &         0.853 &  \bftab 0.935 \\
      &     & 3 & 28     & 0.09 &         0.909 &         0.793 &         0.846 &  \bftab 0.991 &         0.934 &         0.506 &         0.821 &         0.852 &         0.887 \\
      &     & 4 & 88     & 0.39 &  \bftab 0.962 &         0.938 &         0.946 &         0.954 &         0.938 &         0.946 &         0.949 &         0.953 &         0.961 \\
      &     & 5 & 115    & 0.06 &         0.783 &         0.766 &         0.831 &         0.816 &         0.784 &         0.760 &         0.769 &         0.687 &  \bftab 0.856 \\
\midrule
pop-failures & 18  & 0 & 46     & 0.02 &         0.836 &         0.862 &         0.885 &         0.814 &         0.903 &         0.856 &  \bftab 0.914 &         0.573 &         0.903 \\
      &     & 1 & 494    & 0.00 &         0.619 &         0.556 &         0.645 &         0.697 &         0.678 &         0.658 &  \bftab 0.717 &         0.538 &         0.686 \\
\midrule
seeds & 7   & 1 & 70     & 0.03 &         0.919 &         0.947 &         0.932 &         0.950 &         0.941 &         0.955 &  \bftab 0.963 &         0.840 &         0.945 \\
      &     & 2 & 70     & 0.03 &         0.989 &         0.968 &         0.982 &         0.992 &  \bftab 0.998 &         0.988 &         0.993 &         0.989 &         0.987 \\
      &     & 3 & 70     & 0.03 &         0.978 &         0.956 &         0.981 &         0.986 &  \bftab 0.990 &         0.973 &         0.977 &         0.913 &         0.982 \\
\midrule
segment & 19  & 1 & 330    & 0.17 &         0.997 &         0.993 &         0.998 &         0.998 &         0.996 &         0.995 &         0.986 &         0.996 &  \bftab 0.999 \\
      &     & 2 & 330    & 0.17 &         0.999 &         0.998 &         0.999 &         0.997 &         0.997 &         0.995 &         0.992 &  \bftab 1.000 &         0.999 \\
      &     & 3 & 330    & 0.20 &         0.906 &         0.848 &         0.922 &  \bftab 0.947 &         0.918 &         0.899 &         0.859 &         0.933 &         0.944 \\
      &     & 4 & 330    & 0.16 &         0.920 &         0.841 &         0.871 &  \bftab 0.945 &         0.918 &         0.896 &         0.844 &         0.922 &         0.913 \\
      &     & 5 & 330    & 0.26 &         0.945 &         0.902 &         0.920 &  \bftab 0.950 &         0.937 &         0.924 &         0.855 &         0.940 &         0.946 \\
      &     & 6 & 330    & 0.16 &         0.997 &         0.993 &         0.998 &         0.999 &         0.994 &         0.974 &         0.944 &         0.993 &  \bftab 0.999 \\
      &     & 7 & 330    & 0.16 &         0.999 &         0.992 &         0.998 &         0.998 &         0.997 &         0.996 &         0.992 &  \bftab 0.999 &         0.998 \\
\midrule
seismic-bumps & 18  & 0 & 2414   & 0.61 &  \bftab 0.743 &         0.520 &         0.572 &         0.712 &         0.694 &         0.712 &         0.728 &         0.591 &         0.602 \\
      &     & 1 & 170    & 0.52 &         0.346 &         0.428 &         0.415 &  \bftab 0.517 &         0.476 &         0.487 &         0.378 &         0.477 &         0.397 \\
\midrule
sensorless & 48  & 1 & 5319   & 0.00 &         0.980 &         0.974 &         0.987 &         0.947 &         0.969 &         0.897 &         0.909 &         0.947 &  \bftab 0.990 \\
      &     & 2 & 5319   & 0.00 &         0.973 &         0.961 &         0.977 &         0.926 &         0.956 &         0.883 &         0.909 &         0.936 &  \bftab 0.980 \\
      &     & 3 & 5319   & 0.00 &         0.984 &         0.978 &         0.994 &         0.954 &         0.974 &         0.876 &         0.889 &         0.946 &  \bftab 0.995 \\
      &     & 4 & 5319   & 0.00 &         0.985 &         0.977 &         0.992 &         0.938 &         0.966 &         0.893 &         0.877 &         0.883 &  \bftab 0.993 \\
      &     & 5 & 5319   & 0.00 &         0.972 &         0.961 &         0.981 &         0.890 &         0.946 &         0.858 &         0.867 &         0.846 &  \bftab 0.984 \\
      &     & 6 & 5319   & 0.00 &         0.971 &         0.960 &         0.982 &         0.910 &         0.946 &         0.877 &         0.866 &         0.917 &  \bftab 0.985 \\
      &     & 7 & 5319   & 0.00 &         0.996 &         0.989 &         0.997 &         0.997 &         0.986 &         0.941 &         0.932 &         0.991 &  \bftab 0.998 \\
      &     & 8 & 5319   & 0.00 &         0.970 &         0.962 &         0.981 &         0.893 &         0.941 &         0.873 &         0.852 &         0.868 &  \bftab 0.985 \\
      &     & 9 & 5319   & 0.00 &         0.972 &         0.963 &         0.984 &         0.871 &         0.939 &         0.869 &         0.830 &         0.854 &  \bftab 0.987 \\
      &     & 10 & 5319   & 0.00 &  \bftab 0.988 &         0.977 &         0.985 &         0.963 &         0.980 &         0.893 &         0.950 &         0.944 &         0.988 \\
      &     & 11 & 5319   & 0.00 &         0.997 &         0.988 &         0.997 &  \bftab 0.998 &         0.990 &         0.969 &         0.938 &         0.990 &         0.998 \\
\midrule
shuttle & 9   & 1 & 45586  & 0.30 &         0.990 &         0.985 &         0.996 &         0.940 &         0.993 &         0.930 &         0.902 &         0.974 &  \bftab 0.998 \\
      &     & 2 & 50     & 0.37 &         0.892 &         0.866 &         0.931 &         0.958 &         0.962 &         0.843 &         0.812 &         0.930 &  \bftab 0.964 \\
      &     & 3 & 171    & 0.27 &         0.887 &         0.830 &         0.900 &  \bftab 0.947 &         0.944 &         0.827 &         0.783 &         0.835 &         0.924 \\
      &     & 4 & 8903   & 0.31 &         0.988 &         0.986 &         0.992 &         0.986 &         0.987 &         0.973 &         0.956 &         0.989 &  \bftab 0.996 \\
      &     & 5 & 3267   & 0.27 &         0.981 &         0.982 &         0.988 &         0.996 &         0.990 &         0.985 &         0.965 &         0.996 &  \bftab 0.997 \\
      &     & 6 & 10     & 0.31 &         0.930 &         0.878 &         0.871 &         0.548 &         0.946 &         0.754 &         0.673 &         0.656 &  \bftab 0.990 \\
      &     & 7 & 13     & 0.49 &         0.865 &         0.851 &         0.868 &  \bftab 0.964 &         0.926 &         0.468 &         0.737 &         0.871 &         0.924 \\
\midrule
skin & 3   & 1 & 50859  & 0.02 &  \bftab 1.000 &         0.968 &         0.955 &         1.000 &         1.000 &         0.991 &         0.999 &         0.999 &         1.000 \\
      &     & 2 & 194198 & 0.03 &  \bftab 0.997 &         0.839 &         0.909 &         0.889 &         0.426 &         0.892 &         0.916 &         0.948 &         0.963 \\
\midrule
somerville & 6   & 0 & 66     & 0.40 &         0.473 &  \bftab 0.575 &         0.461 &         0.443 &         0.454 &         0.498 &         0.452 &         0.419 &         0.454 \\
      &     & 1 & 77     & 0.46 &         0.631 &         0.567 &         0.629 &         0.602 &  \bftab 0.644 &         0.619 &         0.635 &         0.516 &         0.627 \\
\midrule
sonar & 60  & M & 111    & 0.03 &         0.573 &         0.557 &         0.641 &  \bftab 0.680 &         0.672 &         0.597 &         0.520 &         0.666 &         0.679 \\
      &     & R & 97     & 0.03 &         0.680 &         0.626 &         0.674 &         0.591 &         0.674 &         0.664 &         0.675 &         0.585 &  \bftab 0.705 \\
\midrule
spambase & 57  & 0 & 2788   & 0.81 &         0.794 &         0.576 &         0.646 &         0.798 &         0.756 &  \bftab 0.842 &         0.691 &         0.779 &         0.681 \\
      &     & 1 & 1813   & 0.72 &         0.498 &         0.604 &         0.676 &         0.849 &         0.641 &         0.710 &         0.488 &  \bftab 0.851 &         0.734 \\
\midrule
spectf & 44  & 0 & 55     & 0.14 &         0.842 &         0.820 &  \bftab 0.847 &         0.640 &         0.841 &         0.787 &         0.838 &         0.712 &         0.840 \\
      &     & 1 & 212    & 0.08 &         0.240 &         0.283 &         0.208 &         0.247 &         0.290 &         0.265 &         0.249 &  \bftab 0.326 &         0.249 \\
\midrule
sportsarticles & 59  & objective & 635    & 0.38 &  \bftab 0.848 &         0.605 &         0.638 &         0.835 &         0.836 &         0.839 &         0.829 &         0.814 &         0.651 \\
      &     & subjective & 365    & 0.28 &         0.224 &         0.370 &         0.338 &         0.244 &         0.440 &  \bftab 0.636 &         0.467 &         0.288 &         0.312 \\
\midrule
texture & 40  & 2 & 500    & 0.00 &         0.993 &         0.939 &         0.987 &  \bftab 0.996 &         0.989 &         0.951 &         0.964 &         0.976 &         0.991 \\
      &     & 3 & 500    & 0.00 &         0.999 &         0.988 &         0.999 &  \bftab 1.000 &         0.999 &         0.973 &         0.989 &         0.986 &         0.999 \\
      &     & 4 & 500    & 0.00 &         1.000 &         0.998 &         1.000 &  \bftab 1.000 &         1.000 &         0.995 &         0.998 &         0.999 &         1.000 \\
      &     & 6 & 500    & 0.00 &         0.999 &         0.990 &         0.999 &  \bftab 1.000 &         0.999 &         0.983 &         0.993 &         0.993 &         0.999 \\
      &     & 7 & 500    & 0.00 &  \bftab 1.000 &  \bftab 1.000 &  \bftab 1.000 &  \bftab 1.000 &  \bftab 1.000 &         0.998 &         1.000 &         1.000 &  \bftab 1.000 \\
      &     & 8 & 500    & 0.00 &         0.982 &         0.964 &         0.989 &  \bftab 0.994 &         0.985 &         0.916 &         0.912 &         0.967 &         0.993 \\
      &     & 9 & 500    & 0.00 &         0.993 &         0.963 &         0.994 &  \bftab 0.996 &         0.990 &         0.941 &         0.961 &         0.977 &         0.994 \\
      &     & 10 & 500    & 0.00 &         0.977 &         0.973 &         0.991 &         0.992 &         0.979 &         0.918 &         0.930 &         0.977 &  \bftab 0.992 \\
      &     & 12 & 500    & 0.00 &  \bftab 1.000 &         1.000 &  \bftab 1.000 &  \bftab 1.000 &  \bftab 1.000 &         0.997 &         1.000 &  \bftab 1.000 &  \bftab 1.000 \\
      &     & 13 & 500    & 0.00 &         0.997 &         0.997 &         1.000 &         0.998 &         0.999 &         0.986 &         0.995 &         0.992 &  \bftab 1.000 \\
      &     & 14 & 500    & 0.00 &         0.988 &         0.939 &         0.984 &  \bftab 0.998 &         0.986 &         0.922 &         0.938 &         0.992 &         0.987 \\
\midrule
transfusion & 4   & 0 & 570    & 0.19 &         0.572 &         0.523 &         0.514 &         0.510 &         0.537 &  \bftab 0.597 &         0.574 &         0.571 &         0.562 \\
      &     & 1 & 178    & 0.18 &         0.554 &         0.549 &         0.556 &  \bftab 0.688 &         0.617 &         0.672 &         0.660 &         0.566 &         0.559 \\
\midrule
vehicle & 18  & bus & 218    & 0.15 &         0.969 &         0.922 &         0.967 &  \bftab 0.978 &         0.961 &         0.836 &         0.846 &         0.942 &         0.974 \\
      &     & opel & 212    & 0.10 &         0.768 &         0.665 &         0.696 &  \bftab 0.850 &         0.719 &         0.714 &         0.721 &         0.702 &         0.736 \\
      &     & saab & 217    & 0.09 &         0.759 &         0.719 &         0.717 &  \bftab 0.890 &         0.782 &         0.737 &         0.733 &         0.751 &         0.773 \\
      &     & van & 199    & 0.11 &         0.962 &         0.907 &         0.931 &  \bftab 0.968 &         0.936 &         0.864 &         0.887 &         0.935 &         0.956 \\
\midrule
waveform & 21  & 0 & 1657   & 0.01 &         0.821 &         0.804 &         0.834 &         0.844 &  \bftab 0.852 &         0.847 &         0.848 &         0.775 &         0.831 \\
      &     & 1 & 1647   & 0.01 &         0.885 &         0.864 &         0.893 &         0.895 &         0.906 &  \bftab 0.915 &         0.906 &         0.778 &         0.891 \\
      &     & 2 & 1696   & 0.01 &         0.887 &         0.867 &         0.894 &         0.899 &         0.908 &  \bftab 0.916 &         0.906 &         0.821 &         0.892 \\
\midrule
wdbc & 30  & B & 357    & 0.01 &         0.951 &         0.913 &         0.948 &  \bftab 0.966 &         0.953 &         0.957 &         0.953 &         0.904 &         0.957 \\
      &     & M & 212    & 0.01 &         0.672 &         0.745 &         0.786 &         0.664 &         0.785 &  \bftab 0.874 &         0.726 &         0.633 &         0.823 \\
\midrule
wifi & 7   & 1 & 500    & 0.13 &         0.999 &         0.995 &         0.999 &         0.998 &         0.999 &         0.984 &         0.987 &         0.922 &  \bftab 0.999 \\
      &     & 2 & 500    & 0.11 &         0.972 &         0.938 &         0.968 &         0.925 &  \bftab 0.973 &         0.971 &         0.929 &         0.872 &         0.971 \\
      &     & 3 & 500    & 0.14 &         0.992 &         0.982 &         0.992 &         0.992 &         0.993 &         0.993 &         0.992 &         0.919 &  \bftab 0.994 \\
      &     & 4 & 500    & 0.14 &         0.998 &         0.994 &         0.999 &         0.998 &         0.999 &         0.997 &         0.998 &         0.873 &  \bftab 0.999 \\
\midrule
wilt & 5   & n & 4578   & 0.00 &         0.828 &         0.830 &         0.923 &         0.728 &         0.725 &         0.490 &         0.555 &         0.421 &  \bftab 0.940 \\
      &     & w & 261    & 0.01 &         0.933 &         0.870 &         0.927 &  \bftab 0.972 &         0.947 &         0.881 &         0.900 &         0.862 &         0.943 \\
\midrule
wine & 13  & 1 & 59     & 0.08 &         0.990 &         0.970 &         0.994 &         0.983 &         0.995 &         0.980 &         0.916 &         0.852 &  \bftab 0.996 \\
      &     & 2 & 71     & 0.07 &         0.925 &         0.898 &         0.930 &  \bftab 0.952 &         0.945 &         0.933 &         0.933 &         0.860 &         0.940 \\
      &     & 3 & 48     & 0.08 &         0.999 &         0.982 &         0.997 &  \bftab 1.000 &  \bftab 1.000 &         0.989 &         0.997 &         0.805 &         0.997 \\
\midrule
wisconsin & 9   & 2 & 444    & 0.73 &         0.993 &         0.752 &         0.676 &         0.987 &         0.990 &  \bftab 0.995 &         0.994 &         0.970 &         0.864 \\
      &     & 4 & 239    & 0.31 &         0.566 &         0.848 &         0.781 &         0.820 &         0.903 &  \bftab 0.959 &         0.809 &         0.304 &         0.881 \\
\midrule
wpbc & 32  & N & 110    & 0.04 &         0.565 &         0.593 &         0.550 &         0.517 &         0.528 &         0.550 &         0.572 &  \bftab 0.621 &         0.532 \\
      &     & R & 28     & 0.06 &         0.503 &         0.586 &         0.577 &         0.492 &         0.559 &  \bftab 0.602 &         0.543 &         0.508 &         0.565 \\
\midrule
yeast & 8   & CYT & 463    & 0.36 &  \bftab 0.758 &         0.682 &         0.725 &         0.716 &         0.733 &         0.702 &         0.706 &         0.628 &         0.737 \\
      &     & ERL & 5      & 0.47 &         0.839 &         0.816 &         0.847 &         0.705 &         0.771 &         0.782 &         0.704 &         0.445 &  \bftab 0.867 \\
      &     & EXC & 35     & 0.44 &         0.869 &         0.904 &         0.925 &         0.879 &         0.863 &  \bftab 0.935 &         0.921 &         0.675 &         0.906 \\
      &     & ME1 & 44     & 0.42 &         0.957 &         0.886 &         0.941 &         0.956 &  \bftab 0.962 &         0.867 &         0.830 &         0.730 &         0.958 \\
      &     & ME2 & 51     & 0.41 &         0.732 &         0.736 &         0.737 &         0.706 &         0.719 &  \bftab 0.748 &         0.740 &         0.629 &         0.744 \\
      &     & ME3 & 163    & 0.38 &         0.867 &         0.830 &         0.885 &  \bftab 0.908 &         0.903 &         0.874 &         0.870 &         0.786 &         0.904 \\
      &     & MIT & 244    & 0.38 &         0.721 &         0.681 &         0.728 &  \bftab 0.730 &         0.724 &         0.729 &         0.717 &         0.654 &         0.707 \\
      &     & NUC & 429    & 0.33 &         0.614 &         0.616 &         0.628 &         0.650 &  \bftab 0.658 &         0.632 &         0.632 &         0.612 &         0.635 \\
      &     & POX & 20     & 0.38 &         0.564 &         0.516 &         0.553 &         0.595 &         0.572 &  \bftab 0.661 &         0.551 &         0.640 &         0.585 \\
      &     & VAC & 30     & 0.41 &         0.608 &         0.563 &         0.606 &         0.585 &         0.593 &         0.592 &         0.609 &         0.581 &  \bftab 0.613 \\
\end{longtable}

}

\bibliographystyle{elsarticle-num} 
\bibliography{20200422_oc_params}

~\\[\parskip]

\textbf{Oliver Urs Lenz} holds double MSc degrees in Mathematics from Leiden University and the University of Padova (2011), as well as an MA degree in Linguistics from Leiden University (2012), and has worked as a data scientist and co-founder in Amsterdam and Oslo. He is currently a PhD student at Ghent University.\\[\parskip]

\textbf{Daniel Peralta} is currently a post-doctoral researcher at Ghent University and the VIB (Belgium), within the Data Mining and Modeling for Biomedicine group. He has received the Foundation BBVA Award for Young Computer Science Researchers in 2018. His research interests include data mining, deep learning, biometrics and biological data analysis.\\[\parskip]

\textbf{Chris Cornelis} received the M.Sc. and Ph.D. degrees in computer science from Ghent University, Ghent, Belgium.

He is currently a Postdoctoral Fellow with Ghent University, Belgium, supported by the Odysseus programme of the Science Foundation -- Flanders. His current research interests include fuzzy sets, rough sets, and machine learning.

\end{document}